\documentclass[10pt,letterpaper]{article}
\usepackage[top=0.85in,left=1.5in,footskip=0.75in]{geometry}

\usepackage{amsmath,amssymb}

\usepackage{changepage}

\usepackage[utf8x]{inputenc}

\usepackage{textcomp,marvosym}

\usepackage{cite}

\usepackage{nameref,hyperref}

\usepackage[right]{lineno}

\usepackage{microtype}
\DisableLigatures[f]{encoding = *, family = * }

\usepackage[table]{xcolor}

\usepackage{array}

\newcolumntype{+}{!{\vrule width 2pt}}

\newlength\savedwidth

\usepackage[aboveskip=1pt,labelfont=bf,labelsep=period,justification=raggedright,singlelinecheck=off]{caption}

\makeatletter
\renewcommand{\@biblabel}[1]{\quad#1.}
\makeatother

\date{}

\usepackage{lastpage,fancyhdr,graphicx}
\usepackage{epstopdf}
\fancyheadoffset[L]{2.25in}
\fancyfootoffset[L]{2.25in}
\usepackage[T1]{fontenc}
\usepackage{dsfont}

\PassOptionsToPackage{noplaybutton}{media9}
\usepackage{media9}
\usepackage{comment}
\usepackage{amssymb}
\usepackage{gensymb}
\usepackage{amsmath}
\usepackage{feynmp}
\usepackage[caption=false]{subfig} 
\usepackage{subeqnarray}
\usepackage{mathtools}
\usepackage{relsize}
\usepackage{bm}
\usepackage{graphicx}
\usepackage{bigints}
\usepackage{ifthen}
\usepackage{multirow}
\usepackage{slashbox}
\usepackage{threeparttable}
\usepackage{rotating}
\usepackage{hyperref}
\usepackage[normalem]{ulem}
\usepackage{xcolor}
\usepackage{bm}
\usepackage{bbm}
\usepackage{cleveref}
\crefformat{equation}{Eqn~#2#1#3}
\crefrangeformat{equation}{Eqns~#3#1#4 to~#5#2#6}
\crefmultiformat{equation}{Eqns~#2#1#3}%
{ and~#2#1#3}{, #2#1#3}{ and~#2#1#3}
\crefname{equation}{Eqn}{Eqns}
\Crefname{equation}{Eqn}{Eqns}
\crefname{figure}{Figure}{Figures}
\Crefname{figure}{Figure}{Figures}
\DeclareGraphicsExtensions{.pdf,.png,.eps,.jpg}

\usepackage[draft,footnote,marginclue,innerlayout=inline]{fixme}
\usepackage{todonotes}
\usepackage{marginnote}
\makeatletter
\renewcommand{\@todonotes@drawMarginNoteWithLine}{%
\begin{tikzpicture}[remember picture, overlay, baseline=-0.75ex]%
    \node [coordinate] (inText) {};%
\end{tikzpicture}%
\marginnote[{
    \@todonotes@drawMarginNote%
    \@todonotes@drawLineToLeftMargin%
}]{
    \@todonotes@drawMarginNote%
    \@todonotes@drawLineToRightMargin%
}%
}
\makeatother
\usepackage{setspace}
\fxloadlayouts{marginnote}
\fxsetup{theme=color,mode=multiuser}

\usepackage{tikz}
\usetikzlibrary{intersections,arrows,decorations,decorations.pathmorphing,calc}
\usepackage{pgfplots}
\pgfplotsset{compat=newest}
\pgfplotsset{plot coordinates/math parser=false}

\makeatletter
\let\jnl@style=\rmfamily
\def\ref@jnl#1{{\jnl@style#1}}
\newcommand\aap{\ref@jnl{A\&A}}%
\newcommand\mnras{\ref@jnl{MNRAS}}%
\newcommand\apj{\ref@jnl{ApJ}}%
\catcode`\@=11

\def\beq#1\eeq{\begin{eqnarray}#1\end{eqnarray}}
\def\bseq#1\eseq{\begin{subeqnarray}#1\end{subeqnarray}}
\def\bea#1\eea{\begin{align}#1\end{align}}
\def\bfig#1\efig{\begin{figure}#1\end{figure}}
\def\btikz#1\etikz{\begin{tikzpicture}#1\end{tikzpicture}}
\def\Ds#1#2{{\partial #1}/{\partial #2}}
\def\l{\left}
\def\r{\right}
\newcommand{\Dv}[2]{\frac{\delta #1}{\delta #2}}
\newcommand{\D}[2]{\frac{\partial #1}{\partial #2}}
\newcommand{\dd}[2]{\frac{d #1}{d #2}}
\newcommand{\mvec}[1]{\mathbf{#1}}

\newcommand{\feyneqn}[1]{\begin{fmffile}{#1}\begin{fmfgraph}}
\newcounter{Vnote}

\newcounter{Lnote}

\def\citep{\cite}
\def\citet{\cite}
\renewcommand\footnotemark{}

\newlength{\fw}
\usepackage{array}
\newcolumntype{C}[1]{>{\centering\let\newline\\\arraybackslash\hspace{0pt}}m{#1}}
\newcommand\cell[1]{{\textcolor{red}{#1}}}

\begin{document}
\vspace*{0.2in}

\begin{flushleft}
{\Large
\textbf\newline{
Symplectomorphic registration with phase space regularization by
entropy spectrum pathways.}
}%
\newline
\\
Vitaly L. Galinsky\textsuperscript{1,3\dag},
Lawrence R. Frank\textsuperscript{1,2\ddag}
\\
\bigskip
\textbf{1}
Center for Scientific Computation in Imaging,
  University of California at San Diego, La Jolla, CA 92093-0854, USA
\\
\textbf{2}
Center for Functional MRI, University of California at
  San Diego, La Jolla, CA 92093-0677, USA
\\
\textbf{3}
Electrical and Computer Engineering Department,
  University of California at San Diego, La Jolla, CA 92093-0407, USA
\\
\bigskip

\dag vit@ucsd.edu

\ddag lfrank@ucsd.edu

\end{flushleft}

\section*{Abstract}
The ability to register image data to a common coordinate system is a
critical feature of virtually all imaging studies that
require  multiple subject analysis,
combining single subject data from multiple
modalities, or both.  However, in spite of the
abundance of literature on the subject and the existence of
several variants of registration algorithms, their practical
utility remains
problematic, as commonly acknowledged even by developers of these
methods because the complexity of the problem
has resisted a general, flexible, and robust theoretical and
computational framework.

To address this issue, we present
a new registration method that is
similar in spirit to the current state-of-the-art technique
of diffeomorphic mapping, but is more general and flexible. The
method utilizes a Hamiltonian formalism and constructs registration as
a sequence of symplectomorphic maps in
conjunction with a novel phase space regularization based on
the powerful entropy spectrum pathways (ESP)
framework.

The main advantage of the ESP
regularized symplectomorphic approach versus the standard
approach of coordinates-only diffeomorphic mapping lies in use of
a common metric that remains valid even
with image dependent regularization.
Moreover, the fusion of the
Hamiltonian framework with the ESP theory goes beyond just providing
an alternative spatially varying smoothing strategy - it provides an
efficient and straightforward way to combine multiple
modalities.

The method is demonstrated on the three different magnetic resonance
imaging (MRI) modalities routinely used for human neuroimaging
applications by mapping between high resolution anatomical
(HRA) volumes, medium resolution diffusion weighted MRI
(DW-MRI) and HRA volumes, and low
resolution functional MRI (fMRI) and HRA volumes.
The typical processing time for high
quality mapping ranges from less than a minute to several minutes on a
modern multi core CPU for typical high resolution
anatomical ($\sim 256^3$ voxels) MRI volumes.

For validation of the framework we developed a panel of deformations
expressed in analytical form that includes deformations based on known
physical processes in MRI that
reproduces various distortions and artifacts typically present in
images collected using these different MRI modalities.  Use
of this panel allows us to quantify repeatability and reproducibility
of our method in comparison to several available alternative
approaches. The panel can be used in future studies especially for
quantitative clinical validation of
different registration approaches.

The registration tool will be available as a part of the
QUEST suite from the UCSD Center for Scientific Computation in Imaging
(CSCI).

%

\section{Introduction}
\label{sec:intro}

Modern imaging systems are increasingly capable of acquiring
data sensitive to a wide range of physical parameters at multiple
resolutions, thus offering greater sensitivity to structural and
dynamical information in complex biological systems.  However, these
technological advancements present the increasingly important
theoretical and computational challenge of how to rigorously and
efficiently combine, or \textit{register}, such data in order to be
able to accurately detect and quantify subtle and complex system
characteristics.


The ability to register image data to a common coordinate system
is a critical feature of virtually all imaging studies that require
quantitative statistical analysis of group populations, as well as for
combining single subject modalities.  Consequently, this subject has
been the focus of a great deal of research.  This has been a focus in
computational neuroanatomy which has motivated the developed of
\textit{diffeomorphic} registrations methods
\citep{pmid15551602,pmid17761438,pmid17354694,pmid18979814,pmid22194239}
for which faster and more efficient algorithms continue to be
developed\citep{pmid26221678, pmid24968094, pmid19709963,
  pmid18979813, pmid23685032}, as well as various regularizations
\citep{pmid24409140, pmid20879371} and additional enhancements such as
local-global mixture, contrast changes, multichannel mapping, etc
\citep{pmid24217008, pmid21197460, pmid22972747}, and the use of
probabilistic diffeomorphic registration methods \citep{pmid25320790,
  pmid20879365}.  These registration advancements are important to
group analyses and the development of standard atlases
\citep{pmid20347998,pmid24579121, pmid15501084, pmid23769915,
  pmid21995026, pmid21276861, pmid17354780} which serve a critical
role in the standardization of studies.  The emergence of diffusion
tensor imaging (DTI) methods and their variants for connectivity
studies required the extension of diffeomorphic registration methods
to accommodate tensor data \citep{pmid23880040, pmid22941943,
  pmid20382233, pmid19694253, pmid21134814, pmid19398016,
  pmid21316463, pmid25433212, pmid25333121, pmid24579120,
  pmid23286046, pmid22156979, pmid21761677, pmid18390342}.  These
methods have had a profound effect on the success of numerous
scientific studies on important clinical issues such as Alzheimer's
and traumatic brain injury \citep{pmid24936424, pmid23333372,
  pmid23322456, pmid20879457, pmid20211269, pmid17999940}, as well as
studies in other organs (cardiac, lungs, etc) \citep{pmid24505703,
  pmid22481815, pmid16093505, pmid15508155, pmid20363173}.  Another
important and even more challenging task is a multi-modal registration
(i.e. registering T1 and T2 images, or T1 and DTI, etc), as the
optimal choice of an appropriate objective function is unknown.
Designing and evaluating a universal algorithm that can fit various
applications (among subjects, multi-modal within-subject, multi-modal
across subjects) is an important problem that needs to be addressed,
as existing approaches do not currently posses such universality (see,
e.g., \citet{pmid23739795} for a comprehensive review).

In spite of the abundance of literature and the existence of several
variants of diffeomorphic algorithms their practical appeal are still
rather limited (possibly due to an interplay of a variety of reasons
-- speed, accuracy, robustness, complexity, repeatability, etc), as
commonly acknowledged even by developers of these registration
methods. For example citing the developer of one of the relatively
broadly used approaches -- Large Deformation Diffeomorphic Metric
Mapping \citep{pmid19398016, pmid24579120, pmid23286046, pmid21761677,
  pmid17999940, pmid21521665} -- ``applications of the LDDMM framework
on volumetric 3D medical images still remain limited for practical
reasons'' \citep{LDDMM}.  Two large and thorough comparison studies
(i.e. \citet{pmid19195496, Ribeiro2015}) also confirm that although
currently available methods are in general able to perform the
registration task with varying degrees of success (although some are
exceedingly slow and some are not particular accurate), the practical
use limitations seem to drive an interest in improvements at least in
terms of speed and accuracy.

The recent review paper \citep{pmid27427472} 
conducted a retrospective analysis of the past two decades
of the field of medical image registration since publication of the
original review \citep{pmid10638851}. It is alarming again that the
main conclusion of this twenty years 
retrospective is that in spite of all the progress in the
field of registration ``the two major problems mentioned in
\citep{pmid10638851} -- validation of registration methods and
translation of these to the clinic -- are major problems still, which
have even been aggravated by the elaboration of registration
methods.''

To address these issues we present in this paper a new method that is
similar in spirit to diffeomorphic mapping, but is more general and
flexible.  The transformation is developed within a Hamiltonian
formalism \citep{vialard:tel-00400379,pmid26643025,pmid19059343} in
which not just the spatial coordinates are considered, but the
entirety of phase space%
, which is a called a
\textit{symplectomorphism}. 
This theoretical construct enables a novel
flexible, accurate, and robust computational method based on a
sequence of energy shell transformations.  The incorporation
of phase space constraints allows us to use the same simple metric
on the space of diffeomorphisms that remains valid even with image dependent
regularization, something that is missing in currently available
methods.

The generality of the Hamiltonian framework facilitates the inclusion
of powerful prior information for spatially varying regularization in
phase space using our recently developed method of entropy spectrum
pathways (ESP) \citep{Frank:2014pre}.  This is in contrast with the
current state-of-the-art approaches that introduce
regularization as a differential form (almost always with constant
coefficients) acting on the map itself (see
e.g.~\citet{Beg2005,doi:10.1137/140984002,5204344}), that effectively
apply regularization as an additional post processing step, thus
creating additional problems, especially for validation and comparison
between different approaches and even different regularization
techniques.  Even the existence of several techniques for spatially
varying smoothing strategies that have been recently proposed
\citep{pmid25485406,pmid25333122} do not remediate this validation
issue (and this is in addition to being of rather limited practical
utility, possibly adding even more speed--accuracy--complexity issues
than providing solutions).  Generally speaking, the Hamiltonian
approach facilitates validation of different regularizations without
destroying or modifying the metric on the space of diffeomorphisms.

The importance and the main advantage of symplectomorphic approach
versus coordinates only diffeomorphic mapping can be understood from
the fact that use of the same common metric allows quantitative
assessment of differences between registrations as well as evaluation
of performance for different regularization schemes. An existence of
volumetric/surface/line measures allows accurate comparison of
features between subvolumes, surface areas or linear curves.

What is even more important is that this fusion of the Hamiltonian
framework with the ESP theory goes beyond just providing an
alternative spatially varying smoothing strategy. It provides an
efficient and straightforward way to combine multiple modalities, for
use in tractography, structural and functional connectivity,
etc. (although the details of implementation go beyond the subject of
this paper and will be reported elsewhere).  

Our method also incorporates fast, accurate, and flexible spatial
preconditioning using our spherical wave decomposition (SWD)
\citep{swd}.  The SWD approach uses fast FFT--based algorithms to
expand images in spherical wave modes and therefore allows to do image
resampling, scaling, rotating and filtering with the highest possible
order of polynomial accuracy, but at a fraction of a time.

The method is validated on a well characterized numerical phantom and
then demonstrated on a set of the ``standard'' neuro-MRI data
acquisitions (HRA, DTI, rsFMRI) routinely collected at our UCSD Center
for FMRI (CFMRI).  We demonstrate the ability to accurately
co-register the data volumes in computational times significantly
faster and more accurately than current state-of-the-art methods.  
The resulting image volumes also demonstrate previously unobserved image
contrasts that suggest the ability of our method to uncover more
subtle and important structural features in the data.

It is well known that different MRI acquisition schemes and protocols
may include a variety of incompatible distortions and artifacts due
not only to variations of scanner hardware and
pulse sequence designs but also due to intrinsic
variations in individual subject morphology, as well as just due to
simple motions.  Thus validation of a registration method's
ability to disentangle the complex interplay of the acquisition
details with the physical effects producing distortions within any
particular individuals brain is an exceedingly non-trivial problem.
Therefore, in order to facilitate a more
quantitative validation of all these different
conditions we developed a panel of deformations defined analytically
and based on well-known physical effects present in the
different MRI modalities.  The deformations from the panel can be
applied to images of different modalities and acquisition condition
and potentially can be appropriate for quick and robust validation in
clinical settings as well.  This validation approach is somewhat
similar to Gaussian deformations used in \citet{pmid15896998}, but our
panel includes deformations that can be attributed to a variety of
real physical processes present in different acquisition protocols and
modalities (i.e.~twist, whirl, stretch, etc).

To evaluate the practical aspects of our implementation and to
demonstrate the competitiveness of our approach we compared the
accuracy and speed of phantom registration with several commonly used
registration methods that are often reported as top performers
\citep{pmid19195496} in either speed or accuracy (ANTs Diffeomorphic
Demons, ANTs SyN, FSL FNIRT and AFNI 3dQWarp).  While a variety of
similarity metrics are available, for this paper we used a simple
Root-Mean-Square Deviation (RMSD) as a metric to evaluate
the accuracy of numerical phantom registration and
wall--clock time (that characterizes the human perception of the
passage of time from the start to the completion of a task, referred
to as \textit{time} afterwards) as a practical and intuitive
measure of the algorithms efficiency.

In summary, this paper utilizes a Hamiltonian formalism to develop a
new approach to non-linear flexible image registration. The method
builds a diffeomorphic mapping as a sequence of symplectomorphic maps
with each map embedded in a separate energy shell. The approach adds a
novel phase space regularization based on the
powerful entropy spectrum pathways framework. The framework provides a
unique opportunity to tailor image details into the
regularization scheme by choosing an image derived regularization
kernel.  A spherical wave decomposition is applied as a
powerful preconditioning tool in the position
domain to allow accurate and fast
interpolation, resampling and estimation of fixed shape rotation and
scale.  The result is an efficient and versatile method capable of
fast and accurate registration of a variety of volumetric images of
different modalities and resolutions.

\section{Symplectomorphic mapping}
\label{sec:theory}
We introduce the Hamiltonian function
$\mathcal{H}(\mvec{q},\mvec{p})$ on a fixed Cartesian grid $\mvec{x}$
as
\begin{equation}\label{eq::hamiltonian}
\mathcal{H}(\mvec{q},\mvec{p}) = \frac{1}{2V}\int\l[\mvec{p}^2 +
  \l(I_0(\mvec{x})-I_1(\mvec{q}))\r)^2\r]d\mvec{x}.
\end{equation}
Here $I_0$ and $I_1$ are two multidimensional images defined on the
same fixed Cartesian grid $\mvec{x}$, $V$ is the measure (volume) of
the reference $I_0$ image domain ($V\equiv \int d\mvec{x}$), and
$(\mvec{q}(\mvec{x},t),\mvec{p}(\mvec{x},t))$ is a set of canonical
coordinates, that define a time dependent mapping from Cartesian grid
$\mvec{x}$ to a new curvilinear grid
$\mvec{y}\equiv\mvec{q}(\mvec{x},t)$, such that initially at $t=0$ the
grids are identical, i.e. ($\mvec{q}(\mvec{x},0),\mvec{p}(\mvec{x},0))
\equiv (\mvec{x},0$).

The Hamiltonian \cref{eq::hamiltonian} defines a flow at each location
on a fixed grid through a system of Hamilton's
equations
\begin{align}
\label{eq::flow:q}
\dd{\mvec{q}}{t} &= \Dv{\mathcal{H}}{\mvec{p}} \equiv
\mvec{p}\\
\label{eq::flow:p}
\dd{\mvec{p}}{t} &=-\Dv{\mathcal{H}}{\mvec{q}} \equiv
\l(I_0-I_1\r)\D{I_1}{\mvec{q}}
\end{align}
where $\delta\mathcal{H}/\delta ...$ denotes variational (or
functional) derivative.

The flow defined by \cref{eq::flow:q,eq::flow:p}
is called a \textit{Hamiltonian flow} and takes place in the space
of the coordinates $(\mvec{q},\mvec{p})$, which is called
\textit{phase space}.  Diffeomorphisms in this phase space are
called \textit{Hamiltonian diffeomorphisms} or
\textit{symplectomorphisms} since a phase space is a symplectic
manifold.  Thus symplectomorphisms preserve the symplectic structure
(including the volume) of phase space.  This is a very important
feature that will allow the generation of a shell-like sequence of
transformations suitable for volumetric measurements and
quantifications.

Because the Hamiltonian function \cref{eq::hamiltonian} and the
reference image $I_0$ are defined on a Cartesian grid $\mvec{x}$ we do
not calculate the curvilinear gradient $\Ds{I_1}{\mvec{q}}$
directly. Instead we express $I_1(\mvec{q})$ as a function on
a Cartesian grid $I_1(\mvec{q}(\mvec{x},t))$ and use
the chain rule to evaluate the curvilinear gradient through
a gradient on Cartesian grid $\Ds{I_1}{\mvec{x}}$ and Jacobian
$J\equiv \Ds{\mvec{q}}{\mvec{x}}$ as
$\Ds{I_1}{\mvec{x}}(\Ds{\mvec{q}}{\mvec{x}})^{-1}$.

An evolution of the Jacobian with time can be obtained by differentiating
the position equation (\cref{eq::flow:q}) on a fixed grid, giving a
closed set of equations
\begin{align}
\label{eq::flow1:q}
\dd{\mvec{q}}{t} &= \mvec{p}\\
\label{eq::flow1:p}
\dd{\mvec{p}}{t} &=\l(I_0-I_1\r)\D{I_1}{\mvec{x}}J^{-1}\\
\label{eq::flow1:J}
\dd{J}{t} &=\D{\mvec{p}}{\mvec{x}}
\end{align}
Integrating these equations with initial conditions
$\mvec{q}(\mvec{x},0)=\mvec{x}$, $\mvec{p}(\mvec{x},0)=0$, and
$J(\mvec{x},0) = \mathds{1}$ generates a symplectomorphic
transformation $\mvec{x} \rightarrow \mvec{q}(\mvec{x},t)$.  A new
metric can be defined for the position part $\mvec{q}$ of the
canonical coordinates by introducing the metric tensor $G \equiv
\{g_{ij}\} = {(J^{-1})}^{T} J^{-1}$, where indices $i$ and $j$
correspond to derivatives over $q_i$ and $q_j$ components of the
curvilinear coordinates $\mvec{q}$ such that in Euclidean space
$g_{ij}=\delta_{ij}$ where $\delta_{ij}$ is the Kronecker delta.  The
metric tensor is important for providing accurate measures of line and
surface properties using the curvilinear coordinate system
$\mvec{q}$. For example, a length of a curve parameterized by
${\mvec{x}}(s)$ with a parameter $s$ between zero and one in Cartesian
space can be expressed using the metric tensor and curvilinear mapping
as
\begin{equation}
\int\limits^{1}_{0}\left|\dd{\mvec{x}}{s}\right|ds=
\int\limits^{1}_{0}\sqrt{g_{ij}\dd{q^i}{s}\dd{q^j}{s}}ds,
\end{equation}
where repeated indices $i$ and $j$ represent summation.

To ensure that the transformation is symplectomorphic at every
location on a fixed grid $\mvec{x}$ during numerical integration we set
a small constant $\epsilon$ and impose a requirement that both the
Jacobian and the inverse Jacobian are bounded by this constant, i.e.
\begin{equation}
\label{eq::jacobian:bound}
\epsilon < |J(x,t)| < \epsilon^{-1},
\end{equation}
For the majority of the results presented in the paper a
value of $\epsilon=0.01$ was used.
When the Jacobian becomes sufficiently close to zero the further
integration does not make sense as it will
not be able to guarantee either the symplectomorphic or
diffeomorphic properties of the flow (even
numerical stability of the solution can be compromised).  Therefore,
when the condition of \cref{eq::jacobian:bound} is violated we stop
numerical integration, freeze the flow, and restart the integration
(i.e., setting $t=0$) beginning at a new set of phase space
coordinate $\{\mvec{q}^{(n)}(\mvec{x},0),
\mvec{p}^{(n)}(\mvec{x},0)\}$ where $n$ is the number of restart
times.  Since the Hamiltonian is an operator that describes the
``energy'' of a system, we refer to these $n$ different sets of
initial conditions as \textit{energy shells}.  Each restart of the
integration therefore represents the initiation of a new energy
shell.

The new initial conditions that define the energy shells are
related to the stopping point of the coordinates in the previous
energy shell by the following conditions:
\begin{align}
\label{eq::embedded:ic}
\mvec{q}^{(n)}(\mvec{x},0) &= \mvec{q}^{(n-1)}(\mvec{x},t^{(n)}-t^{(n-1)}),\\
\mvec{p}^{(n)}(\mvec{x},0) &= 0,\\
J^{(n)}(\mvec{x},0) &= \mathds{1}
\end{align}
Repeating this sequence of initial conditions therefore
generates a set of shell-embedded symplectomorphic transformations
such that the total transformation is diffeomorphic with the Jacobian
defined as a product of $J^{(n)}$
\begin{align}
\label{eq::jacobian:shell}
J\l(\mvec{x},t\r) = &J^{(n)}\l(\mvec{x},t-t^{(n)}\r)\cdot
 J^{(n-1)}\l(\mvec{x},t^{(n)}-t^{(n-1)}\r) \cdot 
\ldots \cdot
 J^{0}\l(\mvec{x},t^{(1)}\r)
\end{align}
It is worth noting that this updating
equation for the Jacobian effectively results in an updating of the
metric tensor $G = {(J^{-1})}^{T} J^{-1}$ that
characterizes the local geometry and assures volume preservation.

We would like to emphasize that our use of Hamiltonian framework
provides a major advantage over conventional approaches in both
efficiency and accuracy. For example, similar considerations for
limiting the Jacobian were employed in \citet{pmid18290061} where
Euler equations of viscous flow were used to describe the displacement
field on a fixed grid. The introduction of fixed Eulerian reference
frame resulted in frequent use of costly and inaccurate template
regridding procedure that is completely avoided
by our formulation.

An important practical implementation issue is that the
number of shells $n$ does not have to be introduced in advance and can
be determined based on overall convergence (or even devised from
running time constraints).  In our numerical implementation the shells
were terminated as soon as $I_1 \rightarrow I_0$ convergence condition
\begin{align}\label{eq::convergence}
\int\l[ 
  \vphantom{
    \l(I_0(\mvec{x})-I_1(\mvec{q}^{(n)})\r)^2 
    \l(I_0(\mvec{x})-I_1(\mvec{q}^{(n-1)})\r)^2}
\r. & \l(I_0(\mvec{x})-I_1(\mvec{q}^{(n)})\r)^2
-\l.
\l(I_0(\mvec{x})-I_1(\mvec{q}^{(n-1)})\r)^2\r]d\mvec{x} <0
\end{align}
was not satisfied.

\section{Entropy spectrum pathways as a phase space regularization}
\label{sec:esp}
The form of Hamiltonian function used in \cref{eq::hamiltonian}
assumes only local input from difference between $I_0$ and $I_1$
images to the flow momentum $\mvec{p}$ at every point on the fixed
grid $\mvec{x}$. A more reasonable assumption would be an inclusion of
some information relevant to the structure of $I_0$ and $I_1$
images. One possible (and by far the most straightforward) way to
provide this structure based preconditioning
is the entropy spectrum pathways (ESP) approach
\citep{Frank:2014pre} that takes into account nearest neighbor
coupling between adjacent grid locations.

The ESP approach starts with generating the coupling density
$Q(\mvec{x},\mvec{x}^\prime)$ which can be as simple and trivial as just the
adjacency matrix 
\begin{eqnarray}
Q(\mvec{x},\mvec{x}^\prime) = 
\begin{cases}
1 & \mbox{if $\mvec{x}$ and $\mvec{x}^\prime$ are connected}\\
0 & \mbox{if $\mvec{x}$ and $\mvec{x}^\prime$ are not connected}
\end{cases}
\end{eqnarray}
or may in general include a strength of coupling
through some kind of coupling potentials that may depend on the grid
positions. The ESP approach solves the generalized eigenvalue problem
\begin{equation}\label{eq::esp:eigenproblem}
\lambda\psi(\mvec{x}) = \int Q(\mvec{x},\mvec{x}^\prime) \psi(\mvec{x}^\prime)
d\mvec{x}^\prime,
\end{equation}
finding the largest eigenvalue $\lambda$ and corresponding eigenvector
$\psi(\mvec{x})$ and then constructs the quantity
\begin{equation}\label{eq::esp:tp}
\rho(\mvec{x}^\prime,\mvec{x}) = \frac{Q(\mvec{x},\mvec{x}^\prime)
  \psi(\mvec{x}^\prime)}{\lambda\psi(\mvec{x})}
\end{equation}
calling it the transition probability density for transition between
grid locations $\mvec{x}$ and $\mvec{x}^\prime$. The square of the
eigenvector $\psi(\mvec{x})$ is called the equilibrium probability
$\mu(\mvec{x})$ in the sense that it represents
the stationary solution that satisfies the stationary point condition
\begin{equation}\label{eq::esp:ep}
\mu(\mvec{x}^\prime) = \int
\rho(\mvec{x}^\prime,\mvec{x})
\mu(\mvec{x})
d \mvec{x}
\end{equation}

\cref{eq::esp:tp} can be included in \cref{eq::hamiltonian} to take
into account nonlocal effects and provide a way of regularization
by defining a non-local Hamiltonian
\begin{align}
\label{eq::hamiltonian:non:local}
\mathcal{H}^{nl}(\mvec{q},\mvec{p}) =& \frac{1}{2V}\int \int
\l[
\delta(\mvec{x},\mvec{x}^\prime) \mvec{p}^2 \r.
+ \l.
\rho(\mvec{x},\mvec{x}^\prime)
\l(I_0(\mvec{x}^\prime)-I_1(\mvec{q}))\r)^2\r]d\mvec{x}d\mvec{x}^\prime,
\end{align}
here $\delta(\mvec{x},\mvec{x}^\prime)$ is Dirac delta function,
$\mvec{q}\equiv\mvec{q}(\mvec{x}^\prime,t)$ and
$\mvec{p}\equiv\mvec{p}(\mvec{x}^\prime,t)$. This
nonlocal expression for the Hamiltonian function produces non-local
Hamilton's equations 
\begin{align}
\label{eq::flowR:q}
\dd{\mvec{q}}{t} &= \mvec{p}\\
\label{eq::flowR:p}
\dd{\mvec{p}}{t} &=\int\l[\rho(\mvec{x},\mvec{x}^\prime)
\l(I_0-I_1\r)\D{I_1}{\mvec{x}}J^{-1}
\r]d\mvec{x}^\prime
\\
\label{eq::flowR:J}
\dd{J}{t} &=\D{\mvec{p}}{\mvec{x}}
\end{align}
where the momentum equation (\cref{eq::flowR:p}) is the
non-local version of \cref{eq::flow1:p} that now includes the
convolution of a local potential (gradient of squared image difference
in our case) with a kernel $\rho(\mvec{x},\mvec{x}^\prime)$ that
depends on the coupling between grid locations.

Alternatively the non-local Hamiltonian function can be specified as
\begin{align}
\label{eq::hamiltonian:non:local1}
\mathcal{H}^{nl}(\mvec{q},\mvec{p}) =& \frac{1}{2V}\int \int
\rho(\mvec{x},\mvec{x}^\prime) 
\l[
\mvec{p}^2 
+ 
\l(I_0(\mvec{x}^\prime)-I_1(\mvec{q}))\r)^2\r]d\mvec{x}d\mvec{x}^\prime,
\end{align}
providing alternative non-local form for the coordinate equation (\cref{eq::flow1:q}) as well
\begin{align}
\label{eq::flowR:qn}
\dd{\mvec{q}}{t} &= 
\int\rho(\mvec{x},\mvec{x}^\prime)\,\mvec{p}\,
d\mvec{x}^\prime
\end{align}


Assuming that the coupling density $Q(\mvec{x},\mvec{x}^\prime)$ does
not depend on position $\mvec{x}$ but depends only on a difference
between them (i.e. $Q(\mvec{x},\mvec{x}^\prime) \equiv
Q(\mvec{x}-\mvec{x}^\prime)$), the ESP scheme can provide a variety of
position independent regularization kernels often used as convolution
filters in image registration \citep{pmid17761438}. As a trivial
example, an eigenvalue problem (\cref{eq::esp:eigenproblem}) for
position independent Gaussian coupling density
$Q(\mvec{x}-\mvec{x}^\prime)=\exp(-(\mvec{x}-\mvec{x}^\prime)^T S
(\mvec{x}-\mvec{x}^\prime)))$ in infinite $n$-dimensional domain has
maximum eigenvalue $\lambda = \sqrt{\pi^n/\det{S}}$ and a trivial
eigenvector $\psi(\mvec{x})=\mathrm{const}$, resulting in the commonly
used Gaussian regularization kernel.  This simple
illustration is merely meant to demonstrate that the commonly used
Gaussian kernel is naturally derived from our very general
procedure.  In practice, more complex coupling schemes can provide
more informative prior information, resulting in more robust warping
schemes.

We would like to emphasize the significant
advantages that ESP regularization provides.  Its general
formulation \citep{Frank:2014pre} is probabilistic in nature and
provides a framework for the incorporation of available information.
In the present context of image registration it naturally provides a
mechanism to incorporate information from either or both of the
$I_0$ and $I_1$ images.  The position dependent coupling naturally
creates image dependent regularization.  Moreover, the ESP
approach can also include any information that is not present in
the images themselves but known \textit{a priori} and related to
images in some quantitative way can be easily included into the
coupling scheme with some sort of linear or nonlinear
parameterization.  We have recently demonstrated this ability
to incorporate multiple priors in ESP coupling in the related
problem of multi-modal parameter estimation \citet{quna}, where the
symplectomorphic registration method of this paper was used for 
registration of multiple modalities.
Additionally, incorporation of the ESP method into the
Hamiltonian formalism provides a
simple and efficient way for introduction of different image matching
terms by modification of the position--based part of either
local or nonlocal Hamiltonian function.  This provides great
flexibility for tailoring the method to specific applications.

\section{Spherical waves decomposition as a position domain preconditioning}
\label{sec:swd}

The set of Hamilton's equations
(\cref{eq::flowR:q,eq::flowR:p,eq::flowR:J}) used in the previous
sections to generate a sequence of energy shell-embedded
symplectomorphic transformations (\cref{eq::jacobian:shell}) requires
equal dimensionality of images $I_0$ and $I_1$. However, in many cases
the images to be registered are of different spatial resolutions so
that some form of interpolation is required.  To provide an
effective way to do position domain resampling, interpolation,
filtering and estimation of best orthogonal transform in a single step
we used the spherical waves decomposition (SWD) approach \citep{swd}.

The SWD approach uses fast algorithms to expand both $I_0$ and $I_1$
images in spherical wave modes
\begin{align}\label{eq::swd}
f_{lmn}^{\{0,1\}} =& \int_0^a\int_0^\pi\int_0^{2\pi}
I_{\{0,1\}}(r,\theta,\phi)R_{nl}(r)
Y_{l}^{m\star}(\theta,\phi) r^2 dr
\sin\theta
d\theta d\phi,
\end{align}
where $Y_{l}^{m\star}(\theta,\phi)$ are the spherical harmonics, and
$R_{nl}(r)$ can be expressed through the spherical Bessel function
\begin{equation}\label{eq::bessels}
R_{ln}(r)=\frac{1}{\sqrt{\mathcal{N}_{ln}}}j_l(k_{ln}r),
\end{equation}
with an appropriate choice of normalization constants
$\mathcal{N}_{ln}$ and the discrete spectrum wave numbers $k_{ln}$
determined by the boundary conditions.  The number of modes
($l,m=0\dots L_{max}$ and $n=1\dots N_{max}$) are determined by the
highest image resolution.  The details of definitions of the spherical
harmonics $Y_{l}^{m}(\theta,\phi)$ and spherical Bessel Functions
$j_l(r)$ can be found in \citet{swd}.  The interpolation and
resampling are then implemented as fast inverse spherical wave
transform
\begin{equation}\label{eq::swd:inv}
I_{\{0,1\}}^{NL}(r,\theta,\phi) =
\sum_{n=1}^{N}\sum_{l=0}^{L}\sum_{m=-l}^{l}
\mathcal{F}_{lmn}f_{lmn}^{\{0,1\}}R_{ln}(r)Y_{l}^{m}(\theta,\phi),
\end{equation}
using appropriate grid locations $(r,\theta,\phi)$ and assigning
$f_{lmn}$ to zeros for modes with $n>N_{max}$ or $l,m>L_{max}$. A
variety of low/band/high pass filters can be used for frequency domain
filter $\mathcal{F}$ following the standard image processing
techniques.

The scale and the amount of rigid rotation between images can be
easily and effectively estimated using the decomposition of
the radial and spherical parts using the partial transforms
\begin{align}\label{eq::swd:inv:rtp}
I_{\{0,1\}}^{N}(r) &=\frac{1}{2\sqrt{\pi}}
\sum_{n=1}^{N}
\frac{1}{\sqrt{\mathcal{N}_{0n}}}
\mathcal{F}_{00n}f_{00n}^{\{0,1\}}j_{0}(k_{0n}r),\\
I_{\{0,1\}}^{L}(\theta,\phi) &=
\sum_{l=0}^{L}
\frac{1}{\sqrt{\mathcal{N}_{l1}}}
\sum_{m=-l}^{l}\mathcal{F}_{lm1}f_{lm1}^{\{0,1\}}
Y_{l}^{m}(\theta,\phi),
\end{align}
and finding the parameters of the similarity transformation (scale $s_r$ and
rotation angles $\theta_r$ and $\phi_r$) by solving the two (one and two
dimensional) minimization problems
\begin{align}\label{eq::rigid}
s_r &= \arg \min_{s_r}
\int\limits_{0}^{R_{max}}
\l[\l(I_{0}^{N}(r)\r)^2-\l(I_{1}^{N}(s_r r)\r)^2\r] dr,\\
(\theta_r,\phi_r) &= \arg
\min_{\theta_r \phi_r}
\int\limits_{0}^{2\pi}
\int\limits_{0}^{\pi}
 \l[\l(I_{0}^{L}(\theta,\phi)\r)^2 - 
\l(I_{1}^{L}(\theta -
   \theta_r,\phi-\phi_r)\r)^2\r] d\theta d\phi,
\end{align}
using small number of modes ($L<L_{max}$ and $N<N_{max}$) for initial
coarse search and increasing them to refine the estimate, thus
avoiding being trapped in local minimums and at the same time creating
computationally efficient approach. Criteria similar to
the considerations about optimal order of SWD transform expressed
in \citet{swd} can be used as a stopping condition
for this scale refinement procedure. A similar scale refinement procedure
was also used for the calculation of the symplectomorphic mapping.

\section{Results and Discussion}
\label{sec:results}

All results shown below were coded in standard C/C++ and
parallelized using POSIX threads.

\begin{figure*}[!tbh]
\centering
\includegraphics[width=1.0\fw]{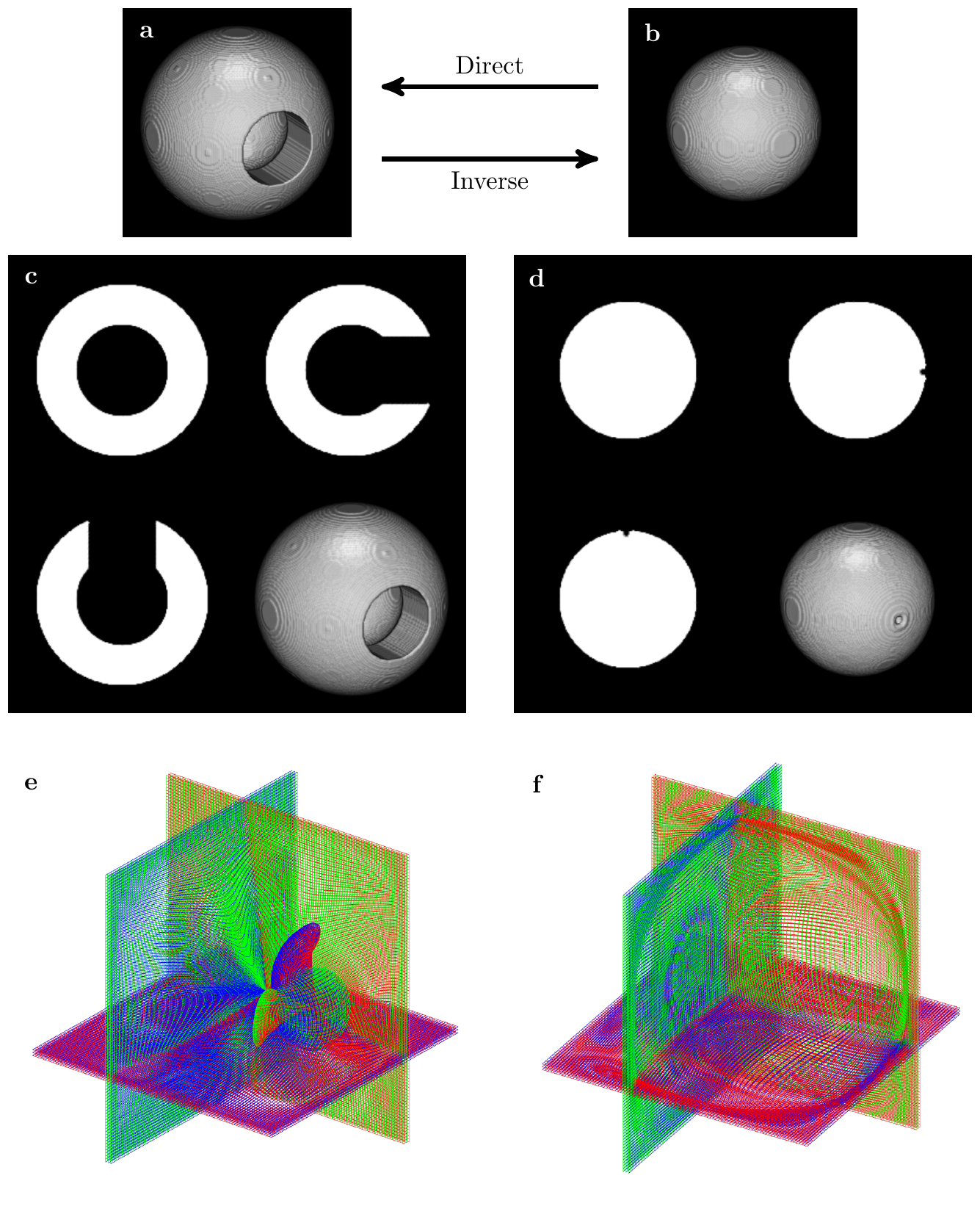}
\caption[]{3D extension of the classical ``toy'' example used for
  benchmarking of diffeomorphic registration: fitting ``circle'' --
  sphere in ({\bf b}) -- to ``C'' -- spherical shell with a hole in
  ({\bf a}). Results of direct ({\bf c}) and inverse ({\bf d}) maps
  obtained in 8 embedded energy shells. Subset of curvilinear grid
  lines plotted for three neighboring layers selected from three
  orthogonal planes for direct ({\bf e}) and inverse ({\bf f})
  maps. The different colors were used to distinguish between the
  anterior-posterior grid lines (blue), the dorsal-ventral grid lines
  (green) and the right-left lateral grid lines (red). Both inverse
  and direct maps were obtained in a single run, the processing time
  for 200x200x200 volumetric datasets was just above 30 seconds on 12
  cores Intel\textregistered\ Core\textsuperscript{TM} i7-4930K CPU \@
  3.40GHz.
\label{fig::toy}
}
\end{figure*}

\subsection{Phantom data registration}
\label{sec:phantom}
To test the approach we first applied it to 3D extension of
a classical ``toy'' example commonly used to show the
performance of non-linear registration approaches --
the registration of the ``C'' shape to
the ''circle'' shape. The original 200x200x200 3D ``C'' and
''circle'' volumes are shown in \cref{fig::toy} ({\bf a}) and ({\bf
  b}). Panels ({\bf c}) and ({\bf d}) show 3D view and central slices
for the forward and inverse maps. Panels ({\bf c}) and ({\bf d}) show
grid lines for a subset of points selected from three groups of
orthogonal planes. The overall performance of our approach seems to be
very good, with nearly perfect forward map of the ``circle'' to the
``C'' shape, and only a slight signature of the original
hole in the inverse map of ``C'' shape to the ``circle''. No SWD
preconditioning was used in this example and simple adjacency type
matrix was used for phase space coupling.

The second stage of our evaluation procedure was the
comparison with some of the existing state-of-the-art non-linear
registration methods commonly used.
As it is clearly beyond the scope of our paper to conduct
a comprehensive evaluation using all possible
variations of nonlinear deformation
algorithms versus all possible variations
of accuracy metrics (such as is
done in evaluation studies, like e.g. \citet{pmid19195496}), we
decided to report here just the simplest and the most straightforward
type of metrics and have restricted our choice of registration tools
to those commonly used at our institution.
Following the recommendation from \citet{pmid19195496} concerning the
speed and accuracy, we processed the phantom registration using
Diffeomorphic Demons (which is reported as one of the fastest) and
symmetric diffeomorphic image registration SyN (reported as one of the
most accurate), both from the Advanced Normalization Tools (ANTs)
package \citep{pmid17659998}, as well as using FNIRT non-linear
registration utility from the FSL \citep{FNIRT} and 3dQwarp non-linear
warping utility from AFNI\citep{pmid8812068}.

As noted above, our quantitative comparison of accuracy and efficiency
are based on two simple and well-understood metrics: The time is used
as a practical measure of efficiency and a simple root mean square
deviation between the reference and the registered image is used as a
measure of accuracy.  The evaluation results are summarized in
\cref{_TABLE_} and demonstrate the enhanced accuracy and efficiency of
our symplectomorphic registration approach in comparison with some
well established non-linear registration techniques for these two
simple metrics and a well-defined standard numerical phantom.

\begin{table*}[!htb]
\renewcommand{\arraystretch}{1.3}
  {\footnotesize
    \caption{Comparison of accuracy and efficiency of phantom
      registration between three commonly used packages (ANTs,
      FSL and AFNI) and our symplectomorphic registration (SYM-REG)
      approach. Simple Root-Mean-Square Deviation (RMSD) between
      reference and registered images are used as an accuracy metric
      and time is reported to characterize an
      efficiency. One of the center slices of registered images is
      also included to provide visual insights into correspondent
      registration artifacts. \label{_TABLE_}}
    \begin{threeparttable}
      {\scriptsize
        \bgroup\def\arraystretch{1.0}%
        \begin{tabular}{l||ccccc}\hline
          & \parbox{0.75cm}{\centering \bfseries ANTs\tnote{a,c}} 
          & \parbox{0.75cm}{\centering \bfseries ANTs\tnote{b,c}}
          & \parbox{0.75cm}{\centering \bfseries FSL\tnote{d}} 
          & \parbox{0.75cm}{\centering \bfseries AFNI\tnote{e}} &
          \bfseries SYM-REG\\
          \hline\hline \bfseries RMSD & 156.4 &
          81.0 & 448.2 & 234.3 & 54.0 \\\hline \bfseries Time & 10min
          15sec & 1h 21min & 12min 43sec & 45min 50sec & 38sec \\\hline \bfseries
            \begin{minipage}{.13\textwidth}
              \hspace*{-8pt}
              \includegraphics[height=20mm,clip,trim=100 100 100 100]
                {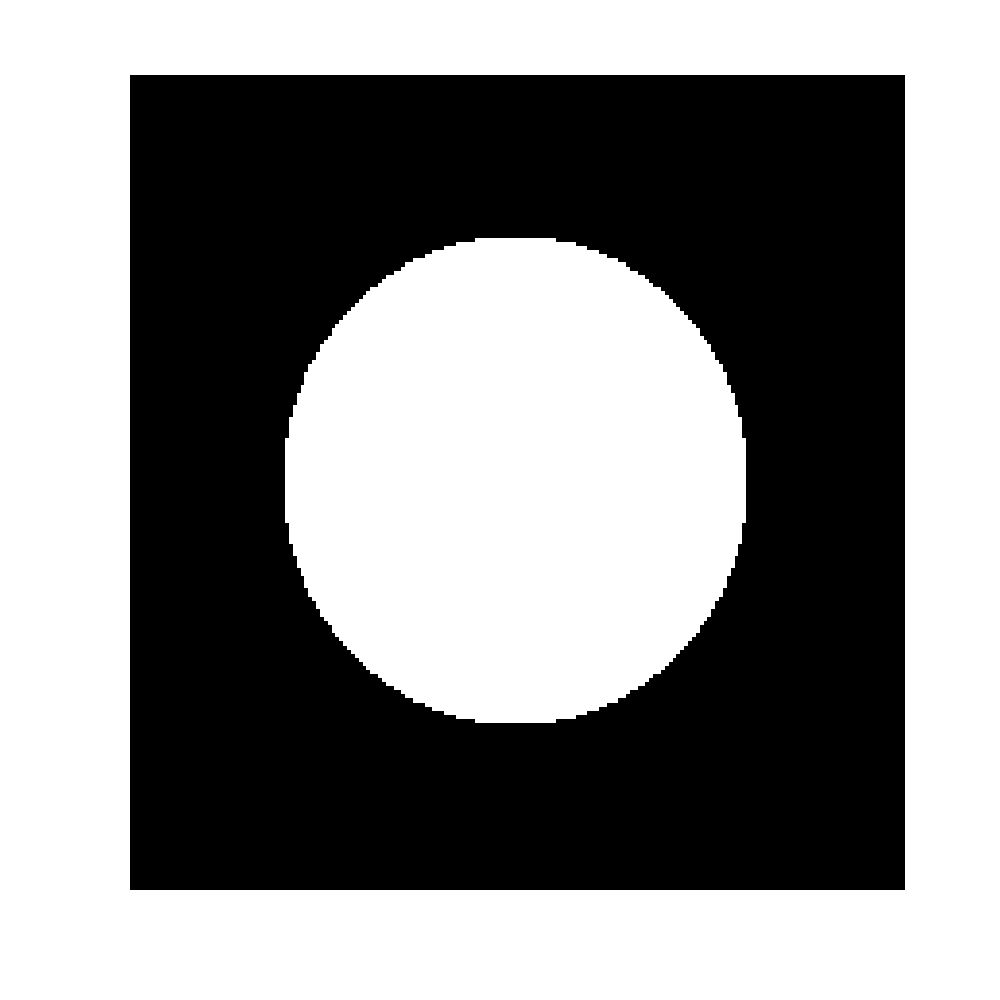}
            \end{minipage}
                        & 
            \begin{minipage}{.13\textwidth}
              \hspace*{-6pt}
              \includegraphics[height=20mm,clip,trim=100 100 100 100]
                {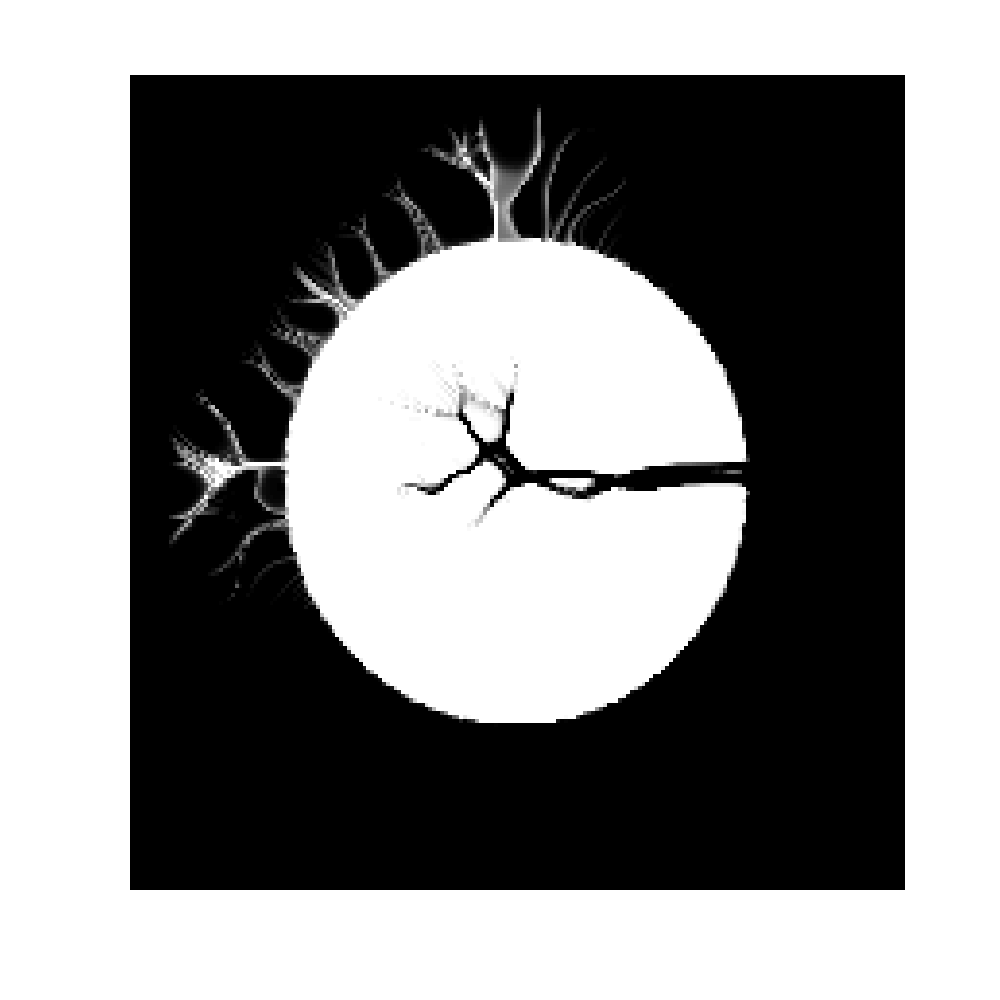}
            \end{minipage}
                        & 
            \begin{minipage}{.13\textwidth}
              \hspace*{-6pt}
              \includegraphics[height=20mm,clip,trim=100 100 100 100]
                {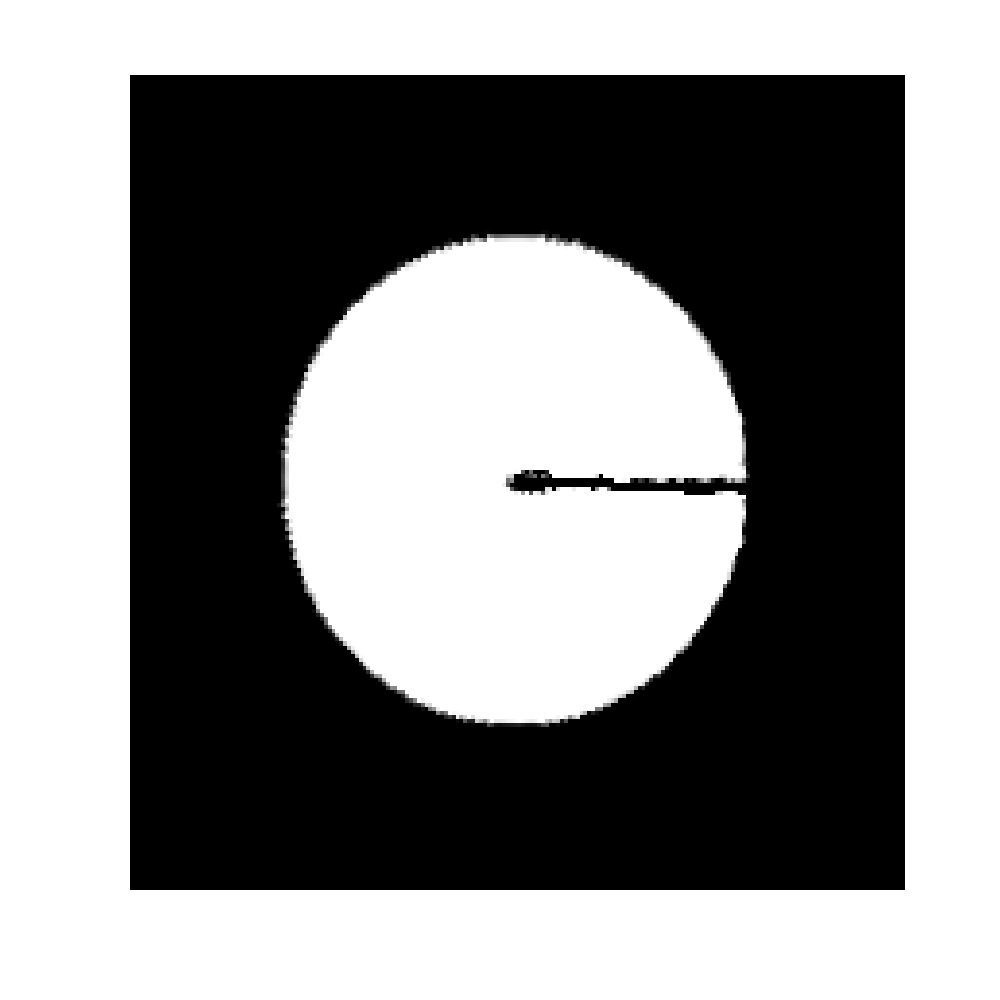}
            \end{minipage}
                        & 
            \begin{minipage}{.13\textwidth}
              \hspace*{-6pt}
              \includegraphics[height=20mm,clip,trim=100 100 100 100]
                {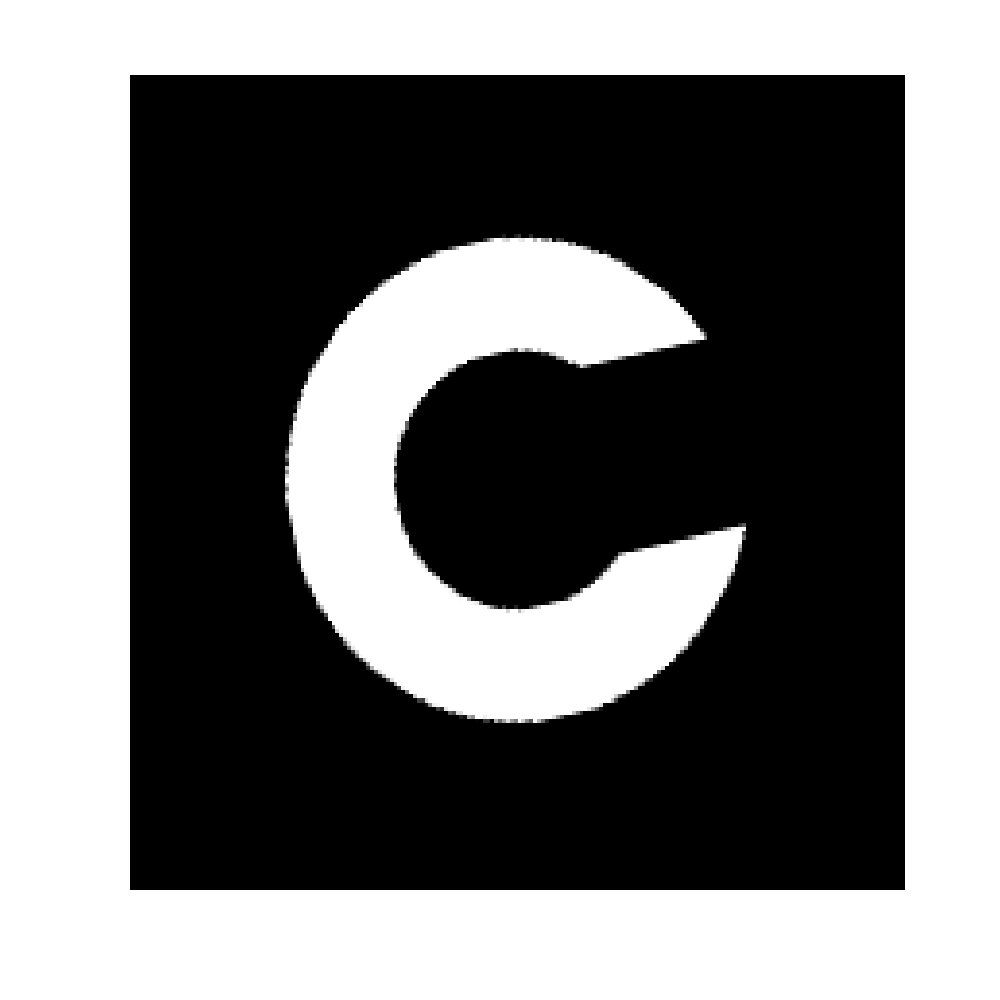}
            \end{minipage}
                        &
            \begin{minipage}{.13\textwidth}
              \hspace*{-6pt}
              \includegraphics[height=20mm,clip,trim=100 100 100 100]
                {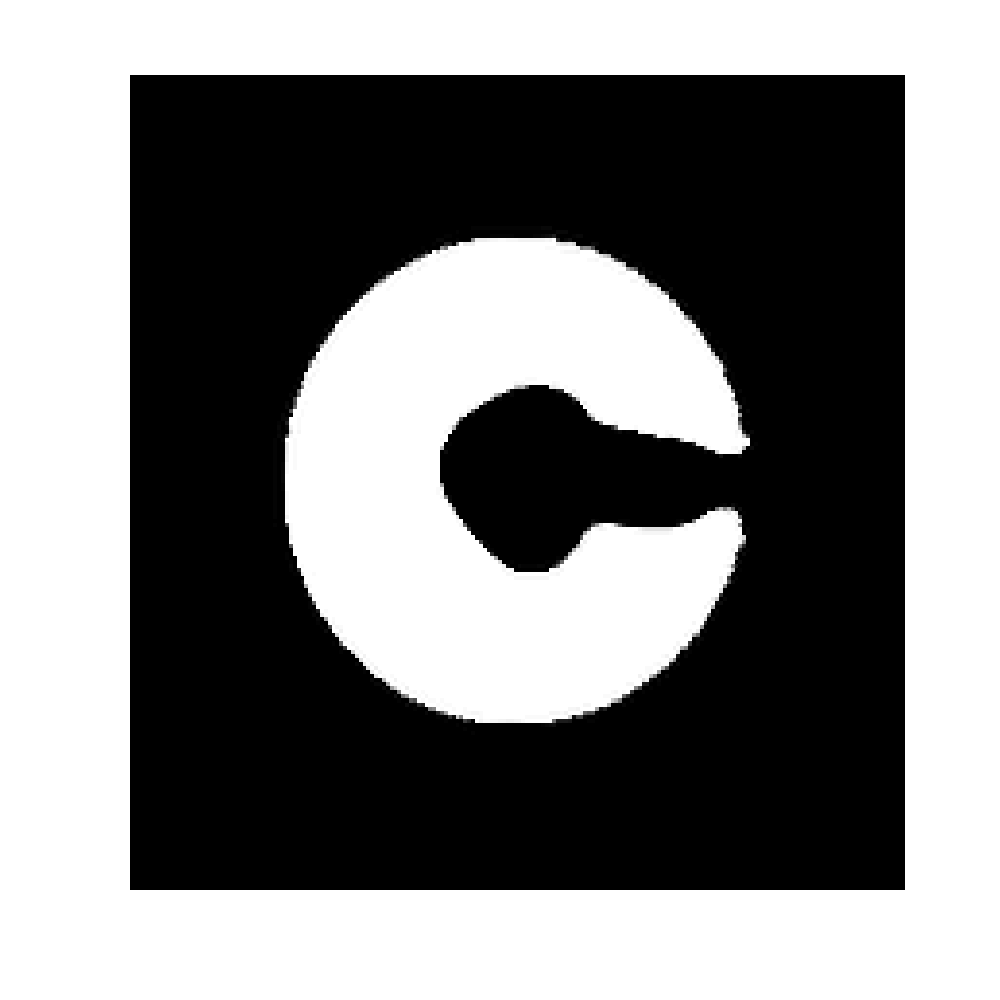}
            \end{minipage}
                        &
            \begin{minipage}{.13\textwidth}
              \hspace*{-6pt}
              \includegraphics[height=20mm,clip,trim=100 100 100 100]
                {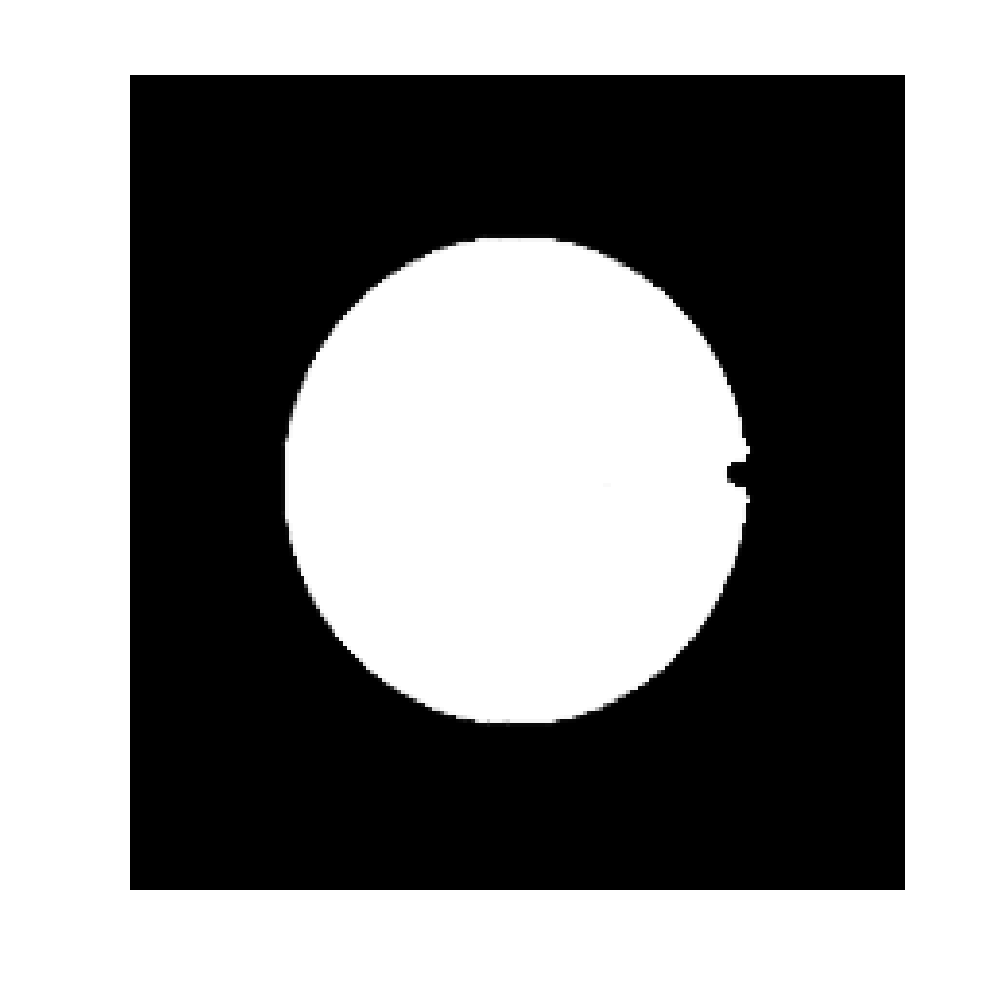}
            \end{minipage}
                        \\\hline
        \end{tabular}
        \egroup
      }
    \end{threeparttable}
    \begin{tablenotes}
    \item {$^a$ {\bfseries GreedyExp} -} We used -t GreedyExp option to produce Diffeomorphic
      Demons style exponential mapping. It was required to increase
      number of iterations using -i 400x160x80 as the default (-i
      10x10x5) choice produced unacceptable results with RMSD=326.
    \item {$^b$ {\bfseries SyN[0.3]} -} This was the best result we were able to achieve
      using native ANTs diffeomorphic registration with -t SyN[0.3]
      parameter. The number of iterations was again increased to
      400x160x80. It may be possible to improve this results by
      further increasing the number of iterations, but we decided that
      going beyond 1h 30min for this simple $200^3$ phantom would not
      justify the improvement.
    \item {$^c$ {\bfseries MSD} -} We used -m MSD (mean square difference) similarity
      model for both of ANTs methods as the rest of the models
      (cross-correlation -- CC, mutual information -- MI, and
      probability mapping -- PR) appeared to be both less accurate and
      more slow for this numerical phantom registration.
    \item {$^d$ {\bfseries FNIRT} -} We used the default (--miter=5,5,5,5) number of
      iterations and it produced the unacceptable results shown
      here. Increasing the number of iterations to 100,100,20,5
      increased the runtime to over an hour with virtually no
      improvement in accuracy.
    \item {$^e$ {\bfseries 3dQWarp} -} We used -penfac 0.02 -workhard options for 3dQwarp to
      allow larger grid deformations and extra iterations to find
      better alignment (the default settings produced unacceptable
      results after 30min with RMSD=418).
    \end{tablenotes}
    \hfill{}
  }
\end{table*}

\subsection{High resolution anatomical MR data registration}
\label{sec:hra}

\subsubsection{Multi-subject registration}

For our first test of the method on actual data, we addressed
the most common usage of registration algorithms: to register
a set
of high resolution anatomical (HRA) images to a common reference
image.  This is a typical multi subject analysis task appearing in a
variety of group studies that involve morphometry based comparison
between different subjects or subject groups.  We utilized HRA data
collected on the 3T GE Discovery MR750 whole body system at the UCSD
Functional MRI (CFMRI) using a 32 channel head coil for ten different
subjects previously collected in a study to determine the effects of
caffeine on the resting state brain activity
\citep{Wong:2013}. However, only high resolution T1 data (all having
290x262x262 voxel resolution) were used for registration test
described in the current subsection.  Further details are available in
\citet{Wong:2013}. Normalized T1 intensities were used as matching
terms $I_0$ and $I_1$ in symplectomorphic registration.

\begin{figure*}[!tbh]
\centering
\includegraphics[width=1.0\fw]{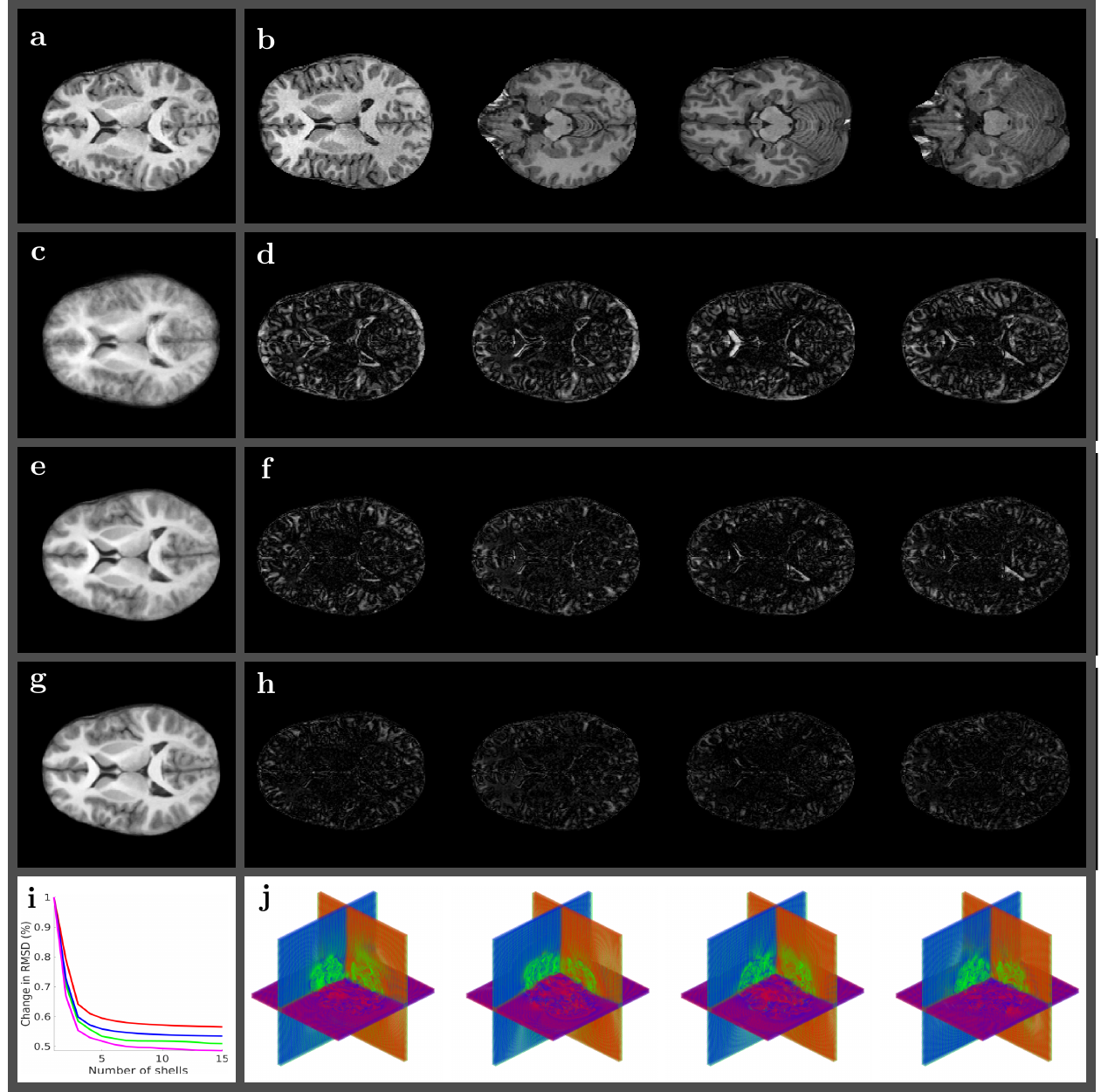}
\caption[]{Results of high resolution anatomical (HRA) mapping to the
  same anatomical reference volume (shown in {\bf a}). ({\bf b})
  Central planes for four volumes out of ten subjects used for
  mapping. ({\bf d}) Residual images of SWD preconditioning (fitted with
  orthogonal transform) for the same four volumes, ({\bf c}) all ten
  volumes averaged. Residual images of symplectomorphic transforms using 5
  ({\bf f}) and 15 ({\bf h}) embedded shells with all ten subject
  averages in ({\bf e}) and ({\bf g}) respectively. 
  ({\bf i}) Convergence plots of the same four subjects as a function of
  a number of shells.
  ({\bf j})
  Illustrative plots of curvilinear grids for the same four subjects,
  using blue color for the anterior-posterior grid lines, green for
  the dorsal-ventral grid lines, and red for the right-left lateral
  grid lines.
\label{fig::anat}
}
\end{figure*}

\cref{fig::anat} shows the collage of images related to this
registration test.  The central plane from anatomical volume used as
reference is shown in ({\bf a}) panel. The same location planes for
randomly selected four volumes out of ten subjects are shown in ({\bf
  b}). Panel ({\bf d}) shows the result of SWD preconditioning step
equivalent to rigidly fitting each volume to reference with orthogonal
transform that includes rotation and uniform scaling for the same four
volumes. Panel ({\bf c}) shows image obtained by averaging of SWD
preconditioned volumes for all ten subjects. The next four panels show
results of symplectomorphic transforms using 5 ({\bf f}) and 15 ({\bf
  h}) embedded energy shells again with correspondent all subject
averages in ({\bf e}) and ({\bf g}) respectively. And finally, panel
({\bf i}) show illustrative plots of curvilinear grids for the same
randomly chosen four subjects.

Overall, as would be expected, the symplectomorphic registration shows
significant improvement over rigidly fitted volumes, with additional
improvement due to increase of a number of energy shells used in
registration.  In general, there is no obvious relationship between
the number of shells and the accuracy, although practically, as the
number of shells is determined by selected limits of Jacobian range
($\epsilon$) (and indirectly can be affected by a selected policy of
time step adjustments), symplectomorphic registration with increased
number of shells may allow to obtain better overall accuracy.  The
total processing time for all ten subject fitting ranges from 15 to 40
minutes based on the selected quality (this is time
measured by running the registration on 12 cores
Intel\textregistered\ Core\textsuperscript{TM} i7-4930K CPU \@
3.40GHz).

We note, while the quality of the fit is measured here by
RMSD, the most advantageous implementation in any particular
clinical or research scenario of course depends on on several
parameters, including the desired quality of the fit, the type of
regularization, the type of coupling in ESP step, aggressiveness of
time step updating, etc.  These trade-offs necessarily depend on the
specific desires of the user in any particular application.

\subsubsection{Synthetic deformation maps}

An important practical aspect of non-linear registration
methods in their application to MRI is their robustness to image
distortions produced in the MRI procedure.  The dependence of
the spatial encoding process in MRI on magnet field gradient linearity
results, in practice, in a wide range of complex non-linear
distortions due to gradient non-linearities induced by both machine
dependent factors (e.g. imperfect gradients) and subject
morphological variations (e.g. susceptibility effects).  To add to
the complexity of this problem, the machine dependent variations can
depend not only on the scanner vendor, but on the scanner software
revision as well.  And the subject morphological variations
certainly differ between subjects. 

Therefore, in order to provide a quantitative
assessment of the symplectomorphic registration using HRA data
under more realistic conditions encountered in practical
applications, we took several T1 brains acquired on different
hardware (Siemens and GE scanners) at different resolutions using
different acquisition sequences and subjected them to
artificially generated distortions designed to mimic some of the
most prominent non-linear distortions common to MRI acquisitions.

\begin{figure*}[!tbh]
\centering
\includegraphics[width=1.0\fw]{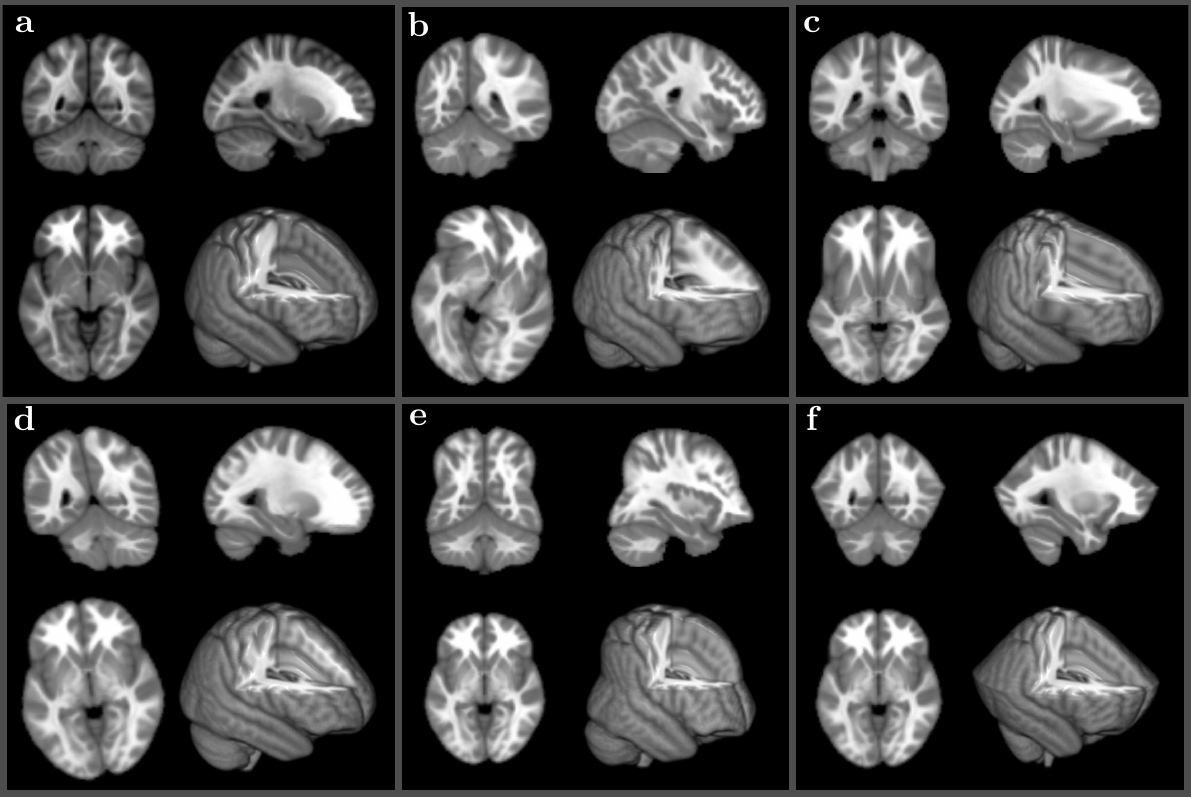}
\caption[]{Different nonlinear warpage types applied to 5 different
  subjects (only one subject is shown for each warpage type): ({\bf
    a}) an original subject; ({\bf b}) differential rotation with the
  amount of rotation proportional to the distance from the center in
  the axial plane (whirl); ({\bf c}) differential stretch in the
  anterior direction; ({\bf d}) differential
  rotation with the amount of rotation proportional to the distance
  from the axial plane in the longitudinal direction (twist); ({\bf
    e}) nonuniform compression in the axial plane proportional to the
  the longitudinal distance; ({\bf f}) nonuniform compression in the
  longitudinal direction relative to the position in the axial plane.
\label{fig::5x5}
}
\end{figure*}

We utilized 5 different high resolution datasets:
(1) MNI152 T1 2mm with 91x109x91 voxels, (2) T1
MPRAGE 1.2mm with 160x200x200 voxels, (3) T1 MPRAGE 1mm with
212x240x256 voxels, (4) T1 1mm with 290x262x262 voxels, and (5) T1 1mm
with 256x176x176 voxels. All subjects were resampled into MNI152 space
(to provide the same accuracy and required workload for all subjects)
and then were distorted several times using five different types of
warpage (see \cref{fig::5x5}). The different types of warpage include
nonlinear or differential rotation, nonlinear stretching, nonlinear
compression, etc, and \cref{fig::5x5} shows one of the subjects with
all five warpage types applied to this subject.

The idea of generation and use of synthetic deformation maps for
validation is not new. A somewhat similar procedure of synthetic
Gaussian 2D deformations was used recently for registration validation
in \citet{pmid15896998}.  The important difference of our deformation
panel is that it includes a variety of more complex deformation modes
observed and frequently present in various MRI acquisition protocols
and modalities.

The 25 warped volumes
were then processed using the two non-linear
registration methods most commonly used at our institution -- ANTs
SyN and AFNI 3dQwarp, in addition to
SYM-REG, and compared the restored volumes with
the original unwarped datasets. The results are summarized in
\cref{_TABLE_HRA_} showing mean values for the RMSD as well as mean
execution times for all subjects (top) and for all warpage types
(bottom). The complete processing took 976, 3414 and 1223 seconds for
SYM-REG, AFNI and ANTs respectively giving 304.6, 358.5 and 584.0 for
the RMSD values with SYM-REG providing both the lowest RMSD
error and the lowest execution time (the default processing options
were used for all packages).

\begin{table*}[!htb]
\renewcommand{\arraystretch}{1.3}
  {\footnotesize
    \caption{Comparison of accuracy and efficiency of HRA registration
      between ANTs SyN, AFNI and our symplectomorphic registration
      (SYM-REG) approach. Five different subjects and five different
      warpage types were used for the comparison. Two subtables
      include mean RMSD values and mean execution times calculated
      across subject (top) and across warpage types (bottom) with the
      best values marked in red. The average of all subjects and
      warpage types gives RMSD of 304.6, 348.5 and 584.0 for SYM-REG,
      AFNI and ANTs respectively with 976, 3414 and 1223 seconds
      required to finish the processing for each method.
      \label{_TABLE_HRA_}}
    \begin{threeparttable}
      {\scriptsize
        \bgroup\def\arraystretch{1.5}%
        \begin{tabular}{l||C{1.55cm}C{1.55cm}C{1.55cm}C{1.55cm}C{1.55cm}}\hline
          \bfseries Method & 
          \multicolumn{5}{c}{
          \parbox{0.5\textwidth}{
          \bfseries Mean RMSD / time (s) across subjects}}\\\hline\hline
          \bfseries AFNI~3dQwarp &402.2/130.8  &363.8/139.4  &371.6/130.0  &316.5/137.8  &338.1/144.8 \\\hline 
          \bfseries ANTs SyN     &578.2/\cell{38.2}  &606.0/47.0  &586.2/43.2  &581.1/60.0  &568.5/56.2  \\\hline 
          \bfseries SYM-REG      &\cell{298.2}/54.8  &\cell{312.4}/\cell{37.0}  &\cell{311.8}/\cell{34.2}  &\cell{300.7}/\cell{33.4}  &\cell{299.8}/\cell{35.8}  \\\hline\hline
          & 
          \multicolumn{5}{c}{
          \parbox{0.5\textwidth}{\hspace*{-10pt}
          \bfseries Mean RMSD / time (s) across warpage types}}\\\hline\hline
          \bfseries AFNI~3dQwarp &162.9/134.0  &373.5/137.2  &482.7/135.6  &404.6/130.0  &368.6/146.0  \\\hline 
          \bfseries ANTs SyN     &140.8/50.6  &705.0/47.4  &602.9/53.6  &781.9/51.0  &689.4/42.0  \\\hline 
          \bfseries SYM-REG      &\cell{87.0}/\cell{38.4}  &\cell{331.6}/\cell{46.8}  &\cell{438.7}/\cell{36.2}  &\cell{351.8}/\cell{38.6}  &\cell{313.9}/\cell{35.2}  \\\hline 
        \end{tabular}
        \egroup
      }
    \end{threeparttable}
    \hfill{}
  }
\end{table*}

\subsection{Diffusion weighed MR image registration}
\label{sec:dwi}

\begin{figure*}[!tbh]
\centering
\includegraphics[width=1.0\fw]{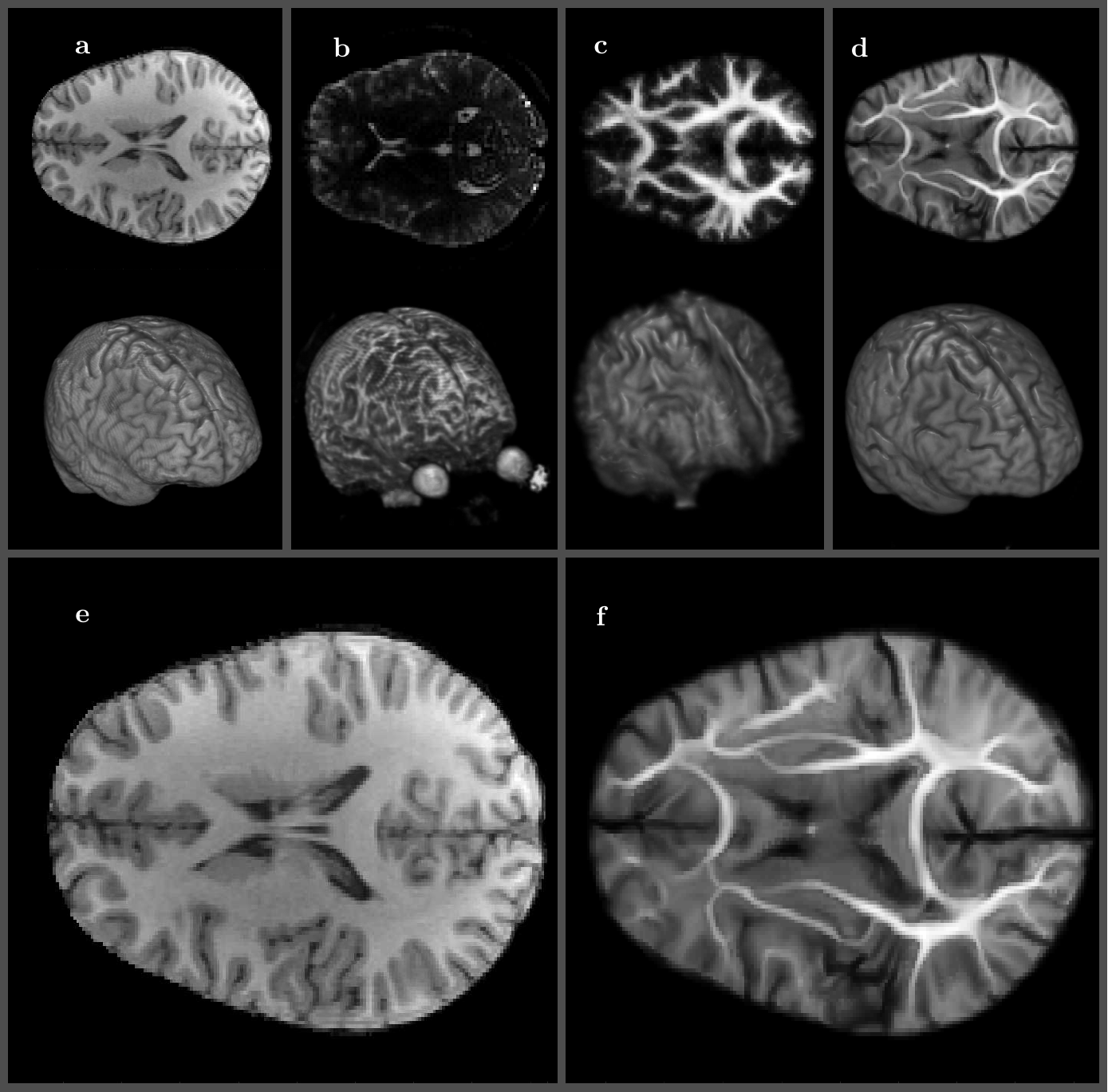}
\caption[]{Medium resolution (100x100x72) diffusion weighted (DWI)
  volume registration to high resolution (168x256x256) T1
  reference. ({\bf a}) Reference T1 MRI image (2D center slice top and
  3D view bottom), ({\bf b}) DWI b0 MRI image, ({\bf c}) equilibrium
  probability DWI image (same resolution as b0 image), ({\bf d}) DWI
  image SWD preconditioned and registered to T1 image (same resolution
  as T1 image). Side by side comparison of reference ({\bf e}) and
  symplectomorphic registration of DWI volume ({\bf f}) demonstrates
  previously unobserved image contrasts additionally differentiating
  white matter and probably emphasizing areas of the highest
  concentration of fibers that suggest the ability of our method to
  uncover more subtle and important structural features in the data.
\label{fig::dwi}
}
\end{figure*}

The second important application we investigated was the
registration of diffusion-weighted images (DWI) to the HRA image of
the same subject.  This is a critical step 
required for collocation of diffusion tractography
based quantities with high resolution anatomical morphometry.
Registration of DWI to HRA is generally a more complicated
problem than registering HRA to HRA because DWI images are typically
acquired with echo-planar imaging (EPI) acquisitions that are not
only prone to more severe non-linear susceptibility distortions than
HRA images but are invariably acquired at lower spatial resolution
than the HRA for the same subject.

The data used for this example were again
collected at the UCSD Center for Functional MRI
(CFMRI) using 3T GE Discovery MR750 whole body system to study the
effects of traumatic brain injuries (TBI).  The HRA T1 volume has
168x256x256 voxel size with 1.2x0.9375x0.9375mm resolution.

A multiband DTI EPI acquisition \citep{Setsompop:2011} using three
simultaneous slice excitations was used to acquire data with three
diffusion sensitizations (at b-values b=1000/2000/3000
s/mm\textsuperscript{2}) for 30, 45 and 65 different diffusion
gradients (respectively) uniformly distributed over a unit sphere.
Several baseline (b=0) images were also recorded.  The data were
reconstructed offline using the CFMRI's multiband reconstruction
routines.  The DWI data has 100x100x72 voxel size with
2mm\textsuperscript{3} resolution.  The normalized HRA T1 intensity
was used for the reference image matching term $I_0$ and the
equilibrium probability map (see \citet{dwi-esp}) was used as a moving
image matching term $I_1$.

\cref{fig::dwi} shows a central slice and a 3D view of the reference
volume ({\bf a}), DWI b=0 volume ({\bf b}), DWI equilibrium
probability volume that has the same resolution as b=0 volume ({\bf
  c}), and the final symplectomorphic registration of the DWI
equilibrium probability volume ({\bf d}). The details about the
equilibrium probability and how it is obtained can be found in
\citet{dwi-esp}. The last two panels ({\bf e}-{\bf f}) show enlarged
side by side comparison of the HRA reference and transformed DWI with
the same resolution.

\begin{figure*}[!tbh]
\centering
\includegraphics[width=1.0\fw]{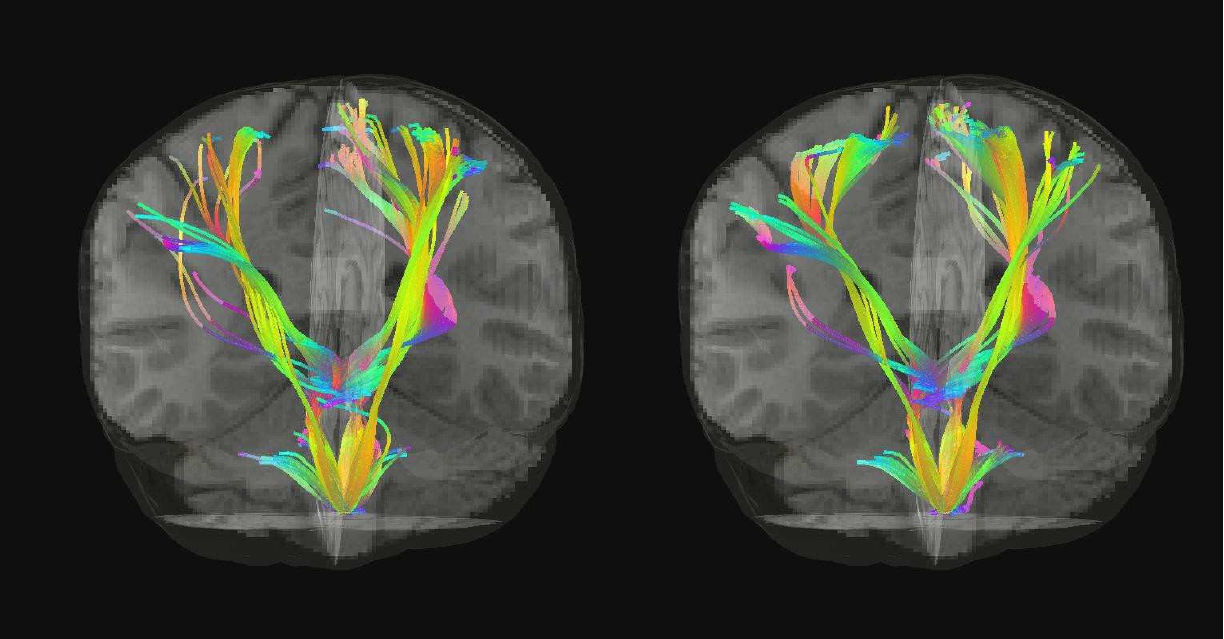}
\caption[]{Two examples of diffusion weighted tractography that
  (left) applied symplectomorphic registration to a set of 
  diffusion weighted tracts, and (right) 
  used
  the GO-ESP \citep{dwi-esp} technique modified by addition
  of HRA dataset through inverse symplectomorphic mapping to the
  diffusion weighted data \citep{quna}.
\label{fig::dwi:trac}
}
\end{figure*}

Two tractography examples
-- one that applies symplectomorphic registration 
to a set of tracts as a postprocessing step, and the other
that includes high resolution symplectomorphic
grid together with the HRA data during the tracking
stage -- are shown in \cref{fig::dwi:trac}.
The figure
shows that the symplectomorphic registration method allows
very accurate localization of diffusion derived tracts with
the high resolution anatomical features.  More details that include a
SYM-REG based tractography implementation as well as multi-modal
estimation in general are not relevant to this paper and are reported
in \citet{quna}.

We would like to mention one important consideration here.  A
considerable amount of work has been spent recently not just on
spatial registration of diffusion imaged volumes, but also on devising
techniques for local reorientation of diffusion tensors that would be
consistent with the new deformed spatial grid [see e.g
  \citet{pmid21134814, pmid23880040, pmid22941943, pmid20382233,
    pmid19398016}].  These methods are both time consuming and an
unnecessary intermediate step from our viewpoint.  An important
feature of our method is that we can directly import the diffeomorphic
maps together with the high resolution data into our diffusion
estimation and tractography technique GO-ESP \citep{dwi-esp},
so that both the estimation of local diffusion properties and
the generation of tracts are performed in the locally warped space
characterized by the spatially dependent metric tensor, thereby
obviating the need to use any \textit{ad hoc} proceed to impose
geometric consistency, which in our method is guaranteed by the
symplectomorphic nature of transformation.  The result is
a method that provides a  fast and effective way
of adding a new level of details to relatively low resolution output
available from diffusion weighted tractography.

\subsection{Registration of functional MR images}
\label{sec:fmri}

The third and final important application we investigated was
the registration of low spatial
resolution functional resting state FMRI (rs-FMRI) data
to HRA data from a single subject.  As in DWI, FMRI data is
typically acquired using EPI acquisitions that, as previously
mentioned, are more prone to non-linear geometric distortions than
HRA acquisitions, and are of lower spatial resolution.

The data used for this test were from the same caffeine
study dataset \citep{Wong:2013} used in \cref{sec:hra}. Registration
of functional rs-FMRI data to anatomical images is required to
establish an accurate localization of activation regions in the high
resolution maps of gray matter.  Only the data collected prior to
caffeine administration were used. Whole brain BOLD resting-state data
were acquired over thirty axial slices using an echo planar imaging
(EPI) sequence (flip angle = $70\degree$, slice thickness=4mm, slice
gap=1mm, FOV=24cm, TE= 30 ms,TR = 1.8 s, matrix size = $64 \times 64
\times 30$).  Further details are available in \citet{Wong:2013}.  All
data were pre-processed using the standard pre-processing analysis
pathway at the CFMRI (as described in \citet{Wong:2013}).  Nuisance
terms were removed from the resting-state BOLD time series through
multiple linear regression.  These nuisance regressors included: i)
linear and quadratic trends, ii) six motion parameters estimated
during image co-registration and their first derivatives, iii)
RETROICOR (2nd order Fourier series) \citep{Glover:2000} and RVHRCOR
\citep{Chang:2009B} physiological noise terms calculated from the
cardiac and respiratory signals, and iv) the mean BOLD signals
calculated from WM and CSF regions and their first respective
derivatives, where these regions were defined using partial volume
thresholds of 0.99 for each tissue type and morphological erosion of
two voxels in each direction to minimize partial voluming with gray
matter.  The normalized HRA T1 intensity was used for the reference
image matching term $I_0$ and the temporal mean of functional
activation was used as a moving image matching term $I_1$.

\begin{figure*}[!tbh]
\centering
\includegraphics[width=1.0\fw]{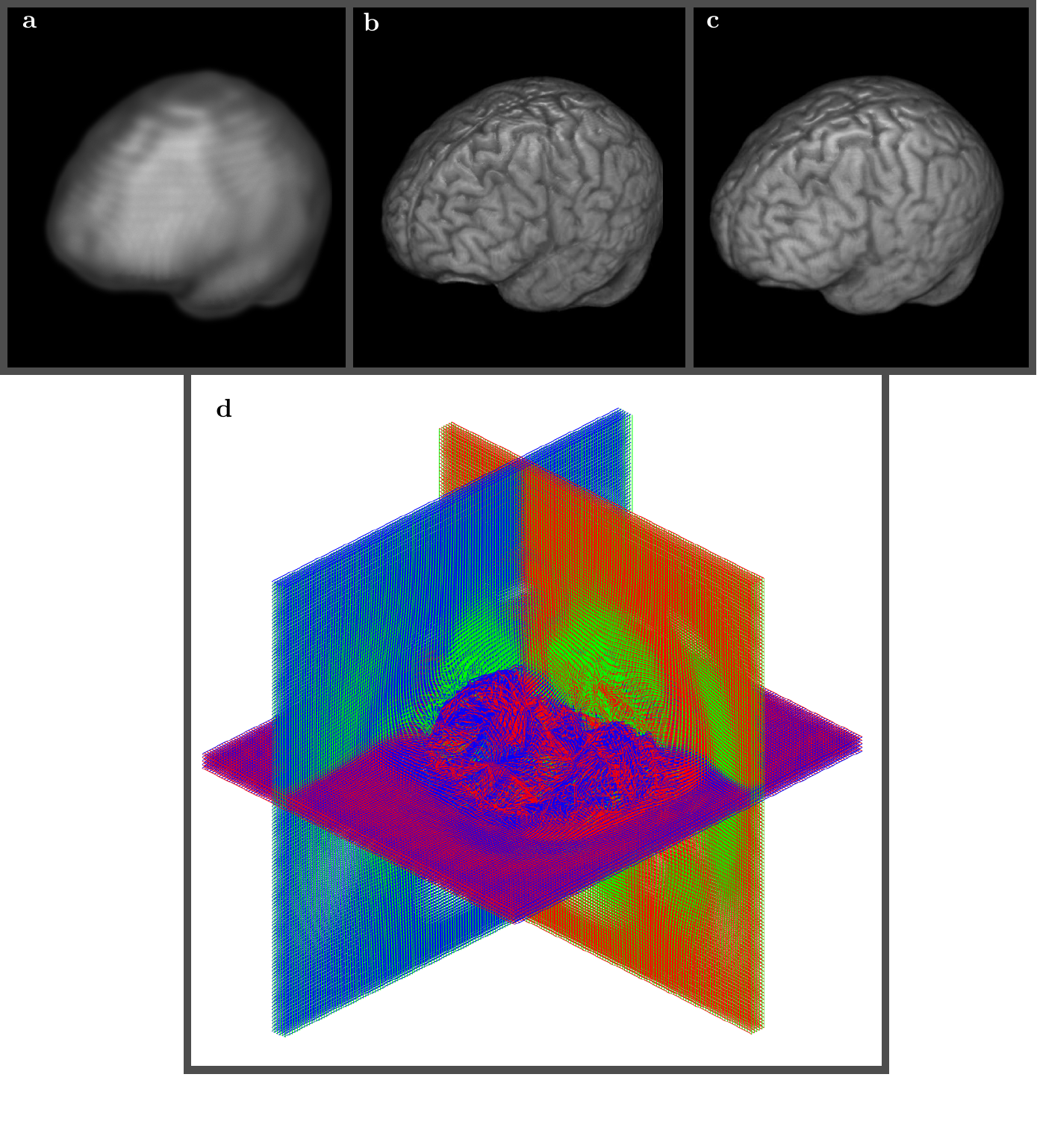}
\caption[]{3D view of low resolution (64x64x30) rs-FMRI volume ({\bf
    a}) vs T1 high resolution (290x262x262) anatomical volume ({\bf
    b}). SWD preconditioned rs-FMRI volume after registration to high
  resolution T1 template ({\bf c}).  The final mapping grid used 30
  shells ({\bf d}) and took about 5 minutes on 12 cores
  Intel\textregistered\ Core\textsuperscript{TM} i7-4930K CPU \@
  3.40GHz. The same color scheme is used for the displacement field
  with blue corresponding to the anterior-posterior grid lines, green
  to the dorsal-ventral grid lines, and red to the right-left lateral
  grid lines.
\label{fig::fmri}
}
\end{figure*}

\cref{fig::fmri} shows side by side comparison for 3D views of rs-FMRI
({\bf a}), T1 ({\bf b}) and rs-FMRI mapped to T1 ({\bf c})
volumes. The processing was carried through 30 energy embedded shells
and required about 5 minutes of waiting time from the start to the
finish, with a subset of the final grid shown in ({\bf d}).  The work
is currently underway to include flexible mapping grids directly to
our rs-FMRI mode detection approach \citep{2016JPhA...49M5001F,fmri-efd}.

\begin{figure*}[!tbh]
\centering
\includegraphics[width=1.0\fw]{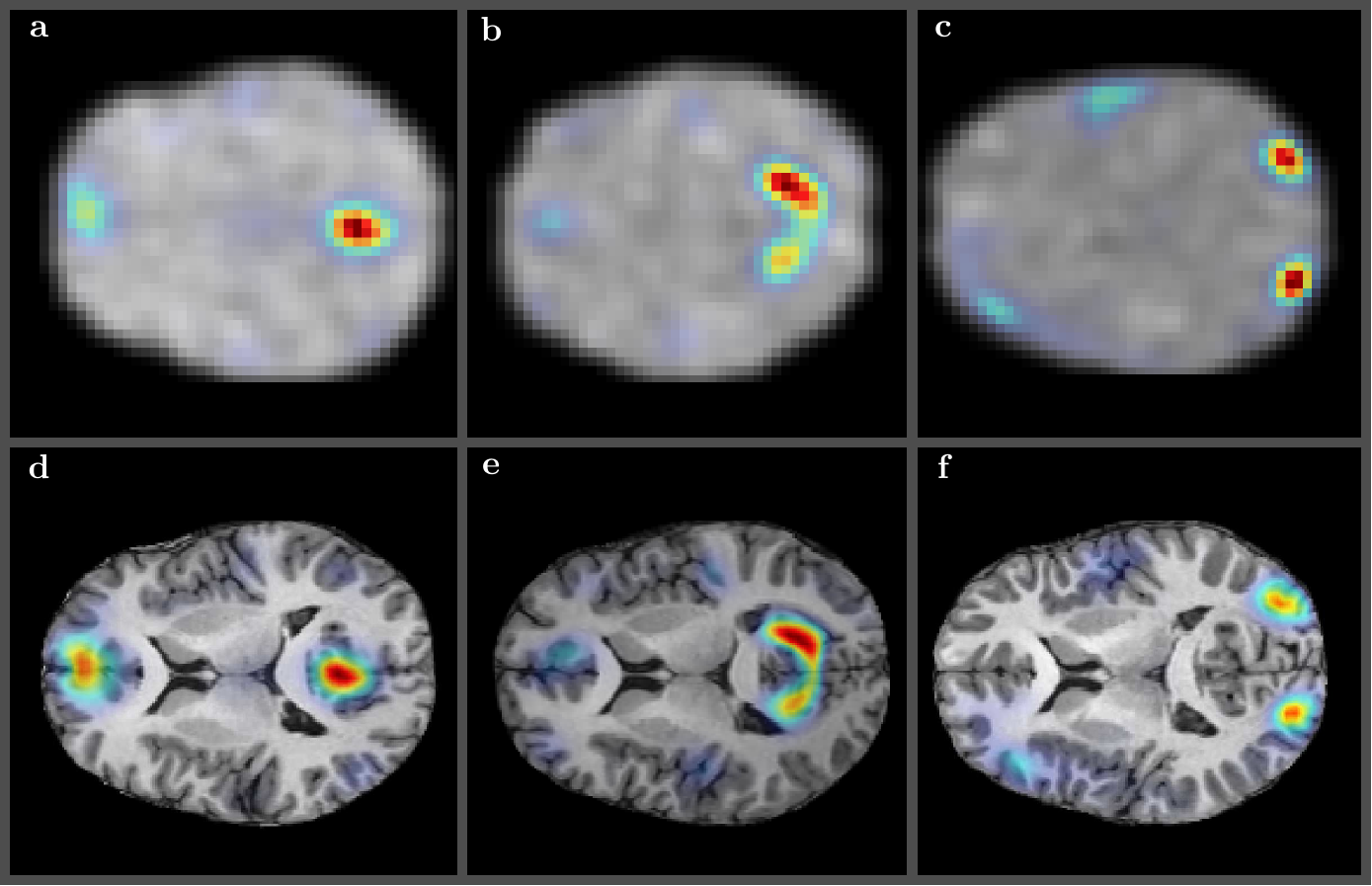}
\caption[]{Several randomly selected resting state modes obtained
  using low resolution (64x64x30) rs-FMRI volume registered to T1 high
  resolution (290x262x262) anatomical volume -- default mode ({\bf a})
  and ({\bf d}), visual lateral ({\bf b}) and ({\bf e}), and visual
  occipital ({\bf c}) and ({\bf e}) -- for some of the subjects from
  \cref{fig::anat}.  The upper panels ({\bf a}, {\bf b}, and {\bf c})
  show the original low resolution rs-FMRI modes. The symplectomorphic
  maps in lower panels ({\bf d}, {\bf e}, and {\bf f}) show accurate
  localizations of functional modes in the appropriate regions of HRA
  volumes.
\label{fig::fmri:modes}
}
\end{figure*}

\cref{fig::fmri:modes} shows comparison of original and registered
images for several of the resting state functional modes obtained
using our entropy field decomposition (EFD) technique
\citep{2016JPhA...49M5001F,fmri-efd}. The default mode ({\bf a}) and ({\bf d}), the
visual lateral ({\bf b}) and ({\bf e}), and the visual occipital ({\bf
  c}) and ({\bf f}) modes are shown for some of the subjects from
\cref{fig::anat}.  The symplectomorphically mapped images overlayed
over correspondent HRA slices ({\bf d}, {\bf e}, and {\bf f}) show
very accurate localization of functional modes in the appropriate
regions of HRA volumes.

\section{Conclusions}
In this paper we have presented a new flexible
multidimensional image registration approach that is based on the
Hamiltonian formalism. The method generates a set of Hamilton's
equations capable of producing a symplectomorphic transformation for
mapping between Cartesian and curvilinear grids
that minimizes some predefined image difference metric.  The final
diffeomorphic mapping is constructed as a multiplicative sequence of
symplectomorphic transforms with gradually diminishing levels of total
energy, thus providing a sequence of energy embedded symplectomorphic
shells.  For demonstration purposes, we used both a simple local
squared difference, as well as a more complicated non-local image
squared difference, as a Hamilton function.

An application of the powerful and versatile ESP approach
\citep{Frank:2014pre} to the phase space domain resulted in a
non-local form of Hamilton's equations. The non-local form represents
an efficient and relatively straight forward way to introduce
regularization that is capable of taking
into account some image specific details or even additional knowledge
based parameterizations.  More generally, the Hamiltonian formalism
allows easy adaptation of custom and possibly more complex forms of
image difference metrics, and at the same time allows the metric in
the diffeomorphism space to be kept the same, which facilitates the
comparison and validation between different techniques and
regularization approaches.

The resolution differences between images as well as rigid shape
alignment is addressed using the preconditioning step based on the SWD
technique \citep{swd}. This efficient volumetric decomposition computes
a set of fast spherical harmonics and spherical Bessel transforms and
is able to produce accurate interpolation, filtering and fitting of
rigid shapes.

One of the major hurdles in the translation of registration
methods to clinical practice \citep{pmid27427472} is the difficulty
in validation of these methods in the face of the exceedingly
complex interplay of the many pulse sequence details, the scanner
hardware, and the anatomical variations that interact with the
magnetic fields. To this end, we have developed a panel
of analytically defined deformations based on known physical
effects present in MR acquisitions.  While these certainly do not
constitute an exhaustive list of possible effects, they are a fair
representation of the major distortions in the major MR image
acquisition schemes.  This deformation panel was used to
validate our method and quantitatively demonstrate its robustness to a
wide range of different types of distortions that regularly plaque MR
studies.  The deformations from the panel can be applied to images
of different modalities and acquisition conditions and potentially can
be appropriate for quick and robust validation in clinical settings.
This validation approach is somewhat similar to 2D Gaussian
deformations used in \citet{pmid15896998}, but our panel includes
deformations that can be attributed to a variety of real physical
processes present in different acquisition protocols and modalities
(i.e.~twist, whirl, stretch, etc).

Overall the symplectomorphic registration approach is both accurate
and fast and is capable of processing of a variety of volumetric
images of different modalities and resolutions.  In the tests reported
in this paper we were able to handle all three of the major neuro-MRI
modalities routinely used for human neuroimaging applications,
including mapping between high-resolution anatomical volumes, medium
resolution diffusion weighted volumes and high-resolution
anatomicals, and low resolution functional MRI images and
high-resolution anatomicals. The typical processing time for
high quality mapping ranges from less than a minute to several minutes
on a modern multi-core CPU for a typical high resolution anatomical
MRI volumes.  The speed, accuracy, and flexibility of this new method
has the potential to play an important
role in the quantitative assessment of neuroimaging data in a wide
range of both basic research and clinical applications.

\section*{Acknowledgments}
The authors thank Dr Alec Wong and Dr Tom Liu at the UCSD CFMRI for
providing the resting state data and Dr Scott Sorg at the VA San Diego
Health Care System for providing the diffusion weighted imaging data.
LRF and VLG were supported by NSF grants DBI-1143389, DBI-1147260,
EF-0850369, PHY-1201238, ACI-1440412, ACI-1550405 and NIH grant R01
MH096100.

\appendix
\section{Individual subject and warpage accuracy and timings} 
\begin{table*}[!htb]
\renewcommand{\arraystretch}{1.3}
  {\footnotesize
    \caption{AFNI~3dQwarp RMSD and time (s)
      for registration of five different subjects and five different 
      warpage types
      \label{_TABLE_AFNI}}
    \begin{threeparttable}
      {\tiny
        \bgroup\def\arraystretch{1.5}%
        \begin{tabular}{l||C{1.65cm}C{1.65cm}C{1.65cm}C{1.65cm}C{1.65cm}}\hline
          \backslashbox[0.25cm]{Subject\hspace*{-15pt}}{\hspace*{-13pt}Warpage\hspace*{-2pt}}
          & 1 & 2 & 3 & 4 & 5\\\hline\hline
          1 & 225.2192 / 129 & 161.0997 / 130 & 185.5213 / 123 & 107.5805 / 132 & 134.8321 / 156 \\\hline 
          2 & 411.3067 / 136 & 380.6455 / 126 & 380.4736 / 123 & 335.5967 / 157 & 359.5755 / 144 \\\hline 
          3 & 520.8890 / 119 & 483.1769 / 167 & 497.8304 / 128 & 452.3812 / 123 & 459.2277 / 141 \\\hline
          4 & 445.6333 / 117 & 417.0812 / 132 & 414.8282 / 125 & 363.8450 / 145 & 381.5591 / 131 \\\hline 
          5 & 408.1614 / 153 & 376.9296 / 142 & 379.3344 / 151 & 323.2617 / 132 & 355.3540 / 152 \\\hline 
        \end{tabular}
        \egroup
      }
    \end{threeparttable}
    \hfill{}
  }
\end{table*}

\begin{table*}[!htb]
\renewcommand{\arraystretch}{1.3}
  {\footnotesize
    \caption{ANTs~SyN RMSD and time (s)
      for registration of five different subjects and five different 
      warpage types
      \label{_TABLE_ANTS}}
    \begin{threeparttable}
      {\tiny
        \bgroup\def\arraystretch{1.5}%
        \begin{tabular}{l||C{1.65cm}C{1.65cm}C{1.65cm}C{1.65cm}C{1.65cm}}\hline
          \backslashbox[0.25cm]{Subject\hspace*{-15pt}}{\hspace*{-13pt}Warpage\hspace*{-2pt}}
          & 1 & 2 & 3 & 4 & 5
          \\\hline\hline
          1 & 144.2440 / 40 & 123.9576 / 46 & 165.7479 / 45 & 129.8583 / 60 & 140.0372 / 62
          \\\hline 
          2 & 685.4163 / 31 & 741.1117 / 48 & 668.2548 / 44 & 714.8960 / 59 & 715.1198 / 55
          \\\hline 
          3 & 595.7664 / 52 & 626.0091 / 46 & 611.8217 / 46 & 583.2656 / 63 & 597.7054 / 61
          \\\hline
          4 & 783.7732 / 34 & 812.4037 / 53 & 784.9238 / 39 & 774.4482 / 69 & 754.1112 / 60
          \\\hline 
          5 & 681.7424 / 34 & 726.7434 / 42 & 700.0624 / 42 & 703.2313 / 49 & 635.3843 / 43
          \\\hline 
        \end{tabular}
        \egroup
      }
    \end{threeparttable}
    \hfill{}
  }
\end{table*}

\begin{table*}[!htb]
\renewcommand{\arraystretch}{1.3}
  {\footnotesize
    \caption{SYM-REG RMSD and time (s)
      for registration of five different subjects and five different 
      warpage types
      \label{_TABLE_SYM_REG}}
    \begin{threeparttable}
      {\tiny
        \bgroup\def\arraystretch{1.5}%
        \begin{tabular}{l||C{1.65cm}C{1.65cm}C{1.65cm}C{1.65cm}C{1.65cm}}\hline
          \backslashbox[0.25cm]{Subject\hspace*{-15pt}}{\hspace*{-13pt}Warpage\hspace*{-2pt}}
          & 1 & 2 & 3 & 4 & 5
          \\\hline\hline
          1 & 81.4292 / 51 &  99.6207 / 50 & 105.7807 / 29 &  76.3845 / 31 &  71.5707 / 31
          \\\hline 
          2 & 331.0654 / 52 & 339.2126 / 29 & 334.6728 / 52 & 320.4282 / 50 & 332.5362 / 51
          \\\hline 
          3 & 420.8638 / 58 & 436.0106 / 30 & 429.0045 / 29 & 482.7290 / 28 & 424.9221 / 36
          \\\hline
          4 & 349.0831 / 50 & 363.1338 / 51 & 362.9932 / 31 & 325.4440 / 28 & 358.1071 / 33
          \\\hline 
          5 & 308.6141 / 63 & 324.2186 / 25 & 326.5479 / 30 & 298.3141 / 30 & 311.8945 / 28
          \\\hline 
        \end{tabular}
        \egroup
      }
    \end{threeparttable}
    \hfill{}
  }
\end{table*}

\Cref{_TABLE_AFNI,_TABLE_ANTS,_TABLE_SYM_REG} provide values of RMSD and wall execution time for all subjects and all warpage types after processing by AFNI~3dQwarp, ANTs SyN, and SYM-REG respectively.


\newpage
\begin{thebibliography}{10}

\bibitem{pmid15551602}
Christensen GE, Rabbitt RD, Miller MI.
\newblock {3{D} brain mapping using a deformable neuroanatomy}.
\newblock Phys Med Biol. 1994;39(3):609--618.

\bibitem{pmid17761438}
Ashburner J.
\newblock {{A} fast diffeomorphic image registration algorithm}.
\newblock Neuroimage. 2007;38(1):95--113.

\bibitem{pmid17354694}
Narayanan R, Fessler JA, Park H, Meyerl CR.
\newblock {{D}iffeomorphic nonlinear transformations: a local parametric
  approach for image registration}.
\newblock Inf Process Med Imaging. 2005;19:174--185.

\bibitem{pmid18979814}
Vercauteren T, Pennec X, Perchant A, Ayache N.
\newblock {{S}ymmetric log-domain diffeomorphic {R}egistration: a demons-based
  approach}.
\newblock Med Image Comput Comput Assist Interv. 2008;11(Pt 1):754--761.

\bibitem{pmid22194239}
Li X, Long X, Laurienti P, Wyatt C.
\newblock {{R}egistration of images with varying topology using embedded maps}.
\newblock IEEE Trans Med Imaging. 2012;31(3):749--765.

\bibitem{pmid26221678}
Zhang M, Fletcher PT.
\newblock {{F}inite-{D}imensional {L}ie {A}lgebras for {F}ast {D}iffeomorphic
  {I}mage {R}egistration}.
\newblock Inf Process Med Imaging. 2015;24:249--259.

\bibitem{pmid24968094}
Gruslys A, Acosta-Cabronero J, Nestor PJ, Williams GB, Ansorge RE.
\newblock {{A} new fast accurate nonlinear medical image registration program
  including surface preserving regularization}.
\newblock IEEE Trans Med Imaging. 2014;33(11):2118--2127.

\bibitem{pmid19709963}
Yeo BT, Sabuncu MR, Vercauteren T, Ayache N, Fischl B, Golland P.
\newblock {{S}pherical demons: fast diffeomorphic landmark-free surface
  registration}.
\newblock IEEE Trans Med Imaging. 2010;29(3):650--668.

\bibitem{pmid18979813}
Yeo BT, Sabuncu M, Vercauteren T, Ayache N, Fischl B, Golland P.
\newblock {{S}pherical demons: fast surface registration}.
\newblock Med Image Comput Comput Assist Interv. 2008;11(Pt 1):745--753.

\bibitem{pmid23685032}
Lorenzi M, Ayache N, Frisoni GB, Pennec X.
\newblock {{L}{C}{C}-{D}emons: a robust and accurate symmetric diffeomorphic
  registration algorithm}.
\newblock Neuroimage. 2013;81:470--483.

\bibitem{pmid24409140}
Tustison NJ, Avants BB.
\newblock {{E}xplicit {B}-spline regularization in diffeomorphic image
  registration}.
\newblock Front Neuroinform. 2013;7:39.

\bibitem{pmid20879371}
Mansi T, Pennec X, Sermesant M, Delingette H, Ayache N.
\newblock {{L}og{D}emons revisited: consistent regularisation and
  incompressibility constraint for soft tissue tracking in medical images}.
\newblock Med Image Comput Comput Assist Interv. 2010;13(Pt 2):652--659.

\bibitem{pmid24217008}
Liu X, Yuan Z, Zhu J, Xu D.
\newblock {{M}edical image registration by combining global and local
  information: a chain-type diffeomorphic demons algorithm}.
\newblock Phys Med Biol. 2013;58(23):8359--8378.

\bibitem{pmid21197460}
Janssens G, Jacques L, Orban~de Xivry J, Geets X, Macq B.
\newblock {{D}iffeomorphic registration of images with variable contrast
  enhancement}.
\newblock Int J Biomed Imaging. 2011;2011:891585.

\bibitem{pmid22972747}
Djamanakova A, Faria AV, Hsu J, Ceritoglu C, Oishi K, Miller MI, et~al.
\newblock {{D}iffeomorphic brain mapping based on {T}1-weighted images:
  improvement of registration accuracy by multichannel mapping}.
\newblock J Magn Reson Imaging. 2013;37(1):76--84.

\bibitem{pmid25320790}
Zhang M, Fletcher PT.
\newblock {{B}ayesian principal geodesic analysis in diffeomorphic image
  registration}.
\newblock Med Image Comput Comput Assist Interv. 2014;17(Pt 3):121--128.

\bibitem{pmid20879365}
Ha L, Prastawa M, Gerig G, Gilmore JH, Silva CT, Joshi S.
\newblock {{I}mage registration driven by combined probabilistic and geometric
  descriptors}.
\newblock Med Image Comput Comput Assist Interv. 2010;13(Pt 2):602--609.

\bibitem{pmid20347998}
Ashburner J, Kloppel S.
\newblock {{M}ultivariate models of inter-subject anatomical variability}.
\newblock Neuroimage. 2011;56(2):422--439.

\bibitem{pmid24579121}
Bieth M, Lombaert H, Reader AJ, Siddiqi K.
\newblock {{A}tlas construction for dynamic (4{D}) {P}{E}{T} using
  diffeomorphic transformations}.
\newblock Med Image Comput Comput Assist Interv. 2013;16(Pt 2):35--42.

\bibitem{pmid15501084}
Joshi S, Davis B, Jomier M, Gerig G.
\newblock {{U}nbiased diffeomorphic atlas construction for computational
  anatomy}.
\newblock Neuroimage. 2004;23 Suppl 1:S151--160.

\bibitem{pmid23769915}
Lim IA, Faria AV, Li X, Hsu JT, Airan RD, Mori S, et~al.
\newblock {{H}uman brain atlas for automated region of interest selection in
  quantitative susceptibility mapping: application to determine iron content in
  deep gray matter structures}.
\newblock Neuroimage. 2013;82:449--469.

\bibitem{pmid21995026}
Dhollander T, Veraart J, Van~Hecke W, Maes F, Sunaert S, Sijbers J, et~al.
\newblock {{F}easibility and advantages of diffusion weighted imaging atlas
  construction in {Q}-space}.
\newblock Med Image Comput Comput Assist Interv. 2011;14(Pt 2):166--173.

\bibitem{pmid21276861}
Oishi K, Mori S, Donohue PK, Ernst T, Anderson L, Buchthal S, et~al.
\newblock {{M}ulti-contrast human neonatal brain atlas: application to normal
  neonate development analysis}.
\newblock Neuroimage. 2011;56(1):8--20.

\bibitem{pmid17354780}
Goodlett C, Davis B, Jean R, Gilmore J, Gerig G.
\newblock {{I}mproved correspondence for {D}{T}{I} population studies via
  unbiased atlas building}.
\newblock Med Image Comput Comput Assist Interv. 2006;9(Pt 2):260--267.

\bibitem{pmid23880040}
Li J, Shi Y, Tran G, Dinov I, Wang DJ, Toga A.
\newblock {{F}ast local trust region technique for diffusion tensor
  registration using exact reorientation and regularization}.
\newblock IEEE Trans Med Imaging. 2014;33(5):1005--1022.

\bibitem{pmid22941943}
Ruthotto L, Kugel H, Olesch J, Fischer B, Modersitzki J, Burger M, et~al.
\newblock {{D}iffeomorphic susceptibility artifact correction of
  diffusion-weighted magnetic resonance images}.
\newblock Phys Med Biol. 2012;57(18):5715--5731.

\bibitem{pmid20382233}
Xue Z, Li H, Guo L, Wong ST.
\newblock {{A} local fast marching-based diffusion tensor image registration
  algorithm by simultaneously considering spatial deformation and tensor
  orientation}.
\newblock Neuroimage. 2010;52(1):119--130.

\bibitem{pmid19694253}
Li H, Xue Z, Guo L, Wong ST.
\newblock {{S}imultaneous consideration of spatial deformation and tensor
  orientation in diffusion tensor image registration using local fast marching
  patterns}.
\newblock Inf Process Med Imaging. 2009;21:63--75.

\bibitem{pmid21134814}
Geng X, Ross TJ, Gu H, Shin W, Zhan W, Chao YP, et~al.
\newblock {{D}iffeomorphic image registration of diffusion {M}{R}{I} using
  spherical harmonics}.
\newblock IEEE Trans Med Imaging. 2011;30(3):747--758.

\bibitem{pmid19398016}
Ceritoglu C, Oishi K, Li X, Chou MC, Younes L, Albert M, et~al.
\newblock {{M}ulti-contrast large deformation diffeomorphic metric mapping for
  diffusion tensor imaging}.
\newblock Neuroimage. 2009;47(2):618--627.

\bibitem{pmid21316463}
Raffelt D, Tournier JD, Fripp J, Crozier S, Connelly A, Salvado O.
\newblock {{S}ymmetric diffeomorphic registration of fibre orientation
  distributions}.
\newblock Neuroimage. 2011;56(3):1171--1180.

\bibitem{pmid25433212}
Irfanoglu MO, Modi P, Nayak A, Hutchinson EB, Sarlls J, Pierpaoli C.
\newblock {{D}{R}-{B}{U}{D}{D}{I} ({D}iffeomorphic {R}egistration for
  {B}lip-{U}p blip-{D}own {D}iffusion {I}maging) method for correcting echo
  planar imaging distortions}.
\newblock Neuroimage. 2015;106:284--299.

\bibitem{pmid25333121}
Irfanoglu MO, Modi P, Nayak A, Knutsen A, Sarlls J, Pierpaoli C.
\newblock {{D}{R}-{B}{U}{D}{D}{I}: diffeomorphic registration for blip up-down
  diffusion imaging}.
\newblock Med Image Comput Comput Assist Interv. 2014;17(Pt 1):218--226.

\bibitem{pmid24579120}
Zhang P, Niethammer M, Shen D, Yap PT.
\newblock {{L}arge deformation diffeomorphic registration of diffusion-weighted
  images with explicit orientation optimization}.
\newblock Med Image Comput Comput Assist Interv. 2013;16(Pt 2):27--34.

\bibitem{pmid23286046}
Zhang P, Niethammer M, Shen D, Yap PT.
\newblock {{L}arge deformation diffeomorphic registration of diffusion-weighted
  images}.
\newblock Med Image Comput Comput Assist Interv. 2012;15(Pt 2):171--178.

\bibitem{pmid22156979}
Du J, Goh A, Qiu A.
\newblock {{D}iffeomorphic metric mapping of high angular resolution diffusion
  imaging based on {R}iemannian structure of orientation distribution
  functions}.
\newblock IEEE Trans Med Imaging. 2012;31(5):1021--1033.

\bibitem{pmid21761677}
Du J, Goh A, Qiu A.
\newblock {{L}arge deformation diffeomorphic metric mapping of orientation
  distribution functions}.
\newblock Inf Process Med Imaging. 2011;22:448--462.

\bibitem{pmid18390342}
Chiang MC, Leow AD, Klunder AD, Dutton RA, Barysheva M, Rose SE, et~al.
\newblock {{F}luid registration of diffusion tensor images using information
  theory}.
\newblock IEEE Trans Med Imaging. 2008;27(4):442--456.

\bibitem{pmid24936424}
Adluru N, Destiche DJ, Lu SY, Doran ST, Birdsill AC, Melah KE, et~al.
\newblock {{W}hite matter microstructure in late middle-age: {E}ffects of
  apolipoprotein {E}4 and parental family history of {A}lzheimer's disease}.
\newblock Neuroimage Clin. 2014;4:730--742.

\bibitem{pmid23333372}
Ota M, Sato N, Matsuo J, Kinoshita Y, Kawamoto Y, Hori H, et~al.
\newblock {{M}ultimodal image analysis of sensorimotor gating in healthy
  women}.
\newblock Brain Res. 2013;1499:61--68.

\bibitem{pmid23322456}
Nakatsuka T, Imabayashi E, Matsuda H, Sakakibara R, Inaoka T, Terada H.
\newblock {{D}iscrimination of dementia with {L}ewy bodies from {A}lzheimer's
  disease using voxel-based morphometry of white matter by statistical
  parametric mapping 8 plus diffeomorphic anatomic registration through
  exponentiated {L}ie algebra}.
\newblock Neuroradiology. 2013;55(5):559--566.

\bibitem{pmid20879457}
Sparks R, Madabhushi A.
\newblock {{N}ovel morphometric based classification via diffeomorphic based
  shape representation using manifold learning}.
\newblock Med Image Comput Comput Assist Interv. 2010;13(Pt 3):658--665.

\bibitem{pmid20211269}
Bossa M, Zacur E, Olmos S.
\newblock {{T}ensor-based morphometry with stationary velocity field
  diffeomorphic registration: application to {A}{D}{N}{I}}.
\newblock Neuroimage. 2010;51(3):956--969.

\bibitem{pmid17999940}
Kim J, Avants B, Patel S, Whyte J, Coslett BH, Pluta J, et~al.
\newblock {{S}tructural consequences of diffuse traumatic brain injury: a large
  deformation tensor-based morphometry study}.
\newblock Neuroimage. 2008;39(3):1014--1026.

\bibitem{pmid24505703}
Zhang Z, Sahn DJ, Song X.
\newblock {{C}ardiac motion estimation by optimizing transmural homogeneity of
  the myofiber strain and its validation with multimodal sequences}.
\newblock Med Image Comput Comput Assist Interv. 2013;16(Pt 1):493--500.

\bibitem{pmid22481815}
Lombaert H, Peyrat JM, Croisille P, Rapacchi S, Fanton L, Cheriet F, et~al.
\newblock {{H}uman atlas of the cardiac fiber architecture: study on a healthy
  population}.
\newblock IEEE Trans Med Imaging. 2012;31(7):1436--1447.

\bibitem{pmid16093505}
Helm P, Beg MF, Miller MI, Winslow RL.
\newblock {{M}easuring and mapping cardiac fiber and laminar architecture using
  diffusion tensor {M}{R} imaging}.
\newblock Ann N Y Acad Sci. 2005;1047:296--307.

\bibitem{pmid15508155}
Beg MF, Helm PA, McVeigh E, Miller MI, Winslow RL.
\newblock {{C}omputational cardiac anatomy using {M}{R}{I}}.
\newblock Magn Reson Med. 2004;52(5):1167--1174.

\bibitem{pmid20363173}
Sofka M, Stewart CV.
\newblock {{L}ocation registration and recognition ({L}{R}{R}) for serial
  analysis of nodules in lung {C}{T} scans}.
\newblock Med Image Anal. 2010;14(3):407--428.

\bibitem{pmid23739795}
Sotiras A, Davatzikos C, Paragios N.
\newblock {{D}eformable medical image registration: a survey}.
\newblock IEEE Trans Med Imaging. 2013;32(7):1153--1190.

\bibitem{pmid21521665}
Risser L, Vialard FX, Wolz R, Murgasova M, Holm DD, Rueckert D, et~al.
\newblock {{S}imultaneous multi-scale registration using large deformation
  diffeomorphic metric mapping}.
\newblock IEEE Trans Med Imaging. 2011;30(10):1746--1759.

\bibitem{LDDMM}
Risser L. Image registration; 2016.
\newblock \url{http://laurent.risser.free.fr/IMPERIAL/menuImperialEng.html}.

\bibitem{pmid19195496}
Klein A, Andersson J, Ardekani BA, Ashburner J, Avants B, Chiang MC, et~al.
\newblock {{E}valuation of 14 nonlinear deformation algorithms applied to human
  brain {M}{R}{I} registration}.
\newblock Neuroimage. 2009;46(3):786--802.

\bibitem{Ribeiro2015}
Ribeiro AS, Nutt DJ, McGonigle J.
\newblock Which Metrics Should Be Used in Non-linear Registration Evaluation?
\newblock In: Navab N, Hornegger J, Wells MW, Frangi FA, editors. Medical Image
  Computing and Computer-Assisted Intervention -- MICCAI 2015: 18th
  International Conference, Munich, Germany, October 5-9, 2015, Proceedings,
  Part II. Cham: Springer International Publishing; 2015. p. 388--395.
\newblock Available from:
  \url{http://dx.doi.org/10.1007/$978-3-319-24571-3_47$}.

\bibitem{pmid27427472}
Viergever MA, Maintz JB, Klein S, Murphy K, Staring M, Pluim JP.
\newblock {{A} survey of medical image registration - under review}.
\newblock Med Image Anal. 2016;33:140--144.

\bibitem{pmid10638851}
Maintz JB, Viergever MA.
\newblock {{A} survey of medical image registration}.
\newblock Med Image Anal. 1998;2(1):1--36.

\bibitem{vialard:tel-00400379}
Vialard FX.
\newblock {Hamiltonian Approach to Shape Spaces in a Diffeomorphic Framework :
  From the Discontinuous Image Matching Problem to a Stochastic Growth Model}
  [Theses].
\newblock {{\'E}cole normale sup{\'e}rieure de Cachan - ENS Cachan}; 2009.
\newblock Available from: \url{https://tel.archives-ouvertes.fr/tel-00400379}.

\bibitem{pmid26643025}
Miller MI, Trouve A, Younes L.
\newblock {{H}amiltonian {S}ystems and {O}ptimal {C}ontrol in {C}omputational
  {A}natomy: 100 {Y}ears {S}ince {D}'{A}rcy {T}hompson}.
\newblock Annu Rev Biomed Eng. 2015;17:447--509.

\bibitem{pmid19059343}
Younes L, Arrate F, Miller MI.
\newblock {{E}volutions equations in computational anatomy}.
\newblock Neuroimage. 2009;45(1 Suppl):40--50.

\bibitem{gee2013information}
Hinkle J, Joshi S.
\newblock {{ID}iff: {I}rrotational {D}iffeomorphisms for {C}omputational
  {A}natomy}.
\newblock In: Gee JC, Joshi S, Pohl KM, Wells WM, Z{\"o}llei L, editors.
  Information Processing in Medical Imaging: 23rd International Conference,
  IPMI 2013, Asilomar, CA, USA, June 28--July 3, 2013, Proceedings. vol.~23 of
  Lecture Notes in Computer Science. Springer Berlin Heidelberg; 2013. p.
  754--765.
\newblock Available from: \url{https://books.google.com/books?id=1Qe7BQAAQBAJ}.

\bibitem{Frank:2014pre}
Frank LR, Galinsky VL.
\newblock Information pathways in a disordered lattice.
\newblock Phys Rev E. 2014;89:032142.
\newblock doi:{10.1103/PhysRevE.89.032142}.

\bibitem{Beg2005}
Beg MF, Miller MI, Trouv{\'e} A, Younes L.
\newblock Computing Large Deformation Metric Mappings via Geodesic Flows of
  Diffeomorphisms.
\newblock International Journal of Computer Vision. 2005;61(2):139--157.
\newblock doi:{10.1023/B:VISI.0000043755.93987.aa}.

\bibitem{doi:10.1137/140984002}
Mang A, Biros G.
\newblock An Inexact Newton--Krylov Algorithm for Constrained Diffeomorphic
  Image Registration.
\newblock SIAM Journal on Imaging Sciences. 2015;8(2):1030--1069.
\newblock doi:{10.1137/140984002}.

\bibitem{5204344}
Hart GL, Zach C, Niethammer M.
\newblock An optimal control approach for deformable registration.
\newblock In: 2009 IEEE Computer Society Conference on Computer Vision and
  Pattern Recognition Workshops; 2009. p. 9--16.

\bibitem{pmid25485406}
Gerig T, Shahim K, Reyes M, Vetter T, Luthi M.
\newblock {{S}patially varying registration using {G}aussian processes}.
\newblock Med Image Comput Comput Assist Interv. 2014;17(Pt 2):413--420.

\bibitem{pmid25333122}
Vialard FX, Risser L.
\newblock {{S}patially-varying metric learning for diffeomorphic image
  registration: a variational framework}.
\newblock Med Image Comput Comput Assist Interv. 2014;17(Pt 1):227--234.

\bibitem{swd}
Galinsky VL, Frank LR.
\newblock {A}utomated {S}egmentation and {S}hape {C}haracterization of
  {V}olumetric {D}ata.
\newblock NeuroImage. 2014;92:156--168.
\newblock doi:{http://dx.doi.org/10.1016/j.neuroimage.2014.01.053}.

\bibitem{pmid15896998}
Rogelj P, Kovacic S.
\newblock {{S}ymmetric image registration}.
\newblock Med Image Anal. 2006;10(3):484--493.

\bibitem{pmid18290061}
Christensen GE, Rabbitt RD, Miller MI.
\newblock {{D}eformable templates using large deformation kinematics}.
\newblock IEEE Trans Image Process. 1996;5(10):1435--1447.

\bibitem{quna}
Galinsky VL, Frank LR.
\newblock {{A} unified theory of neuro-{MRI} data shows scale--free nature of
  connectivity modes}.
\newblock Neural Comput. 2017;0(0):1--27.
\newblock doi:{10.1162/NECO\_a\_00955}.

\bibitem{pmid17659998}
Avants BB, Epstein CL, Grossman M, Gee JC.
\newblock {{S}ymmetric diffeomorphic image registration with cross-correlation:
  evaluating automated labeling of elderly and neurodegenerative brain}.
\newblock Med Image Anal. 2008;12(1):26--41.

\bibitem{FNIRT}
Andersson JLR, Jenkinson M, Smith S. Non-linear registration aka Spatial
  normalisation; 2016.
\newblock FMRIB Technial Report TR07JA2
  \url{http://www.fmrib.ox.ac.uk/analysis/techrep/tr07ja2/tr07ja2.pdf}.

\bibitem{pmid8812068}
Cox RW.
\newblock {{A}{F}{N}{I}: software for analysis and visualization of functional
  magnetic resonance neuroimages}.
\newblock Comput Biomed Res. 1996;29(3):162--173.

\bibitem{Wong:2013}
Wong CW, Olafsson V, Tal O, Liu TT.
\newblock {The amplitude of the resting-state fMRI global signal is related to
  EEG vigilance measures}.
\newblock Neuroimage. 2013;83:983--990.

\bibitem{Setsompop:2011}
Setsompop K, Gagoski BA, Polimeni JR, Witzel T, Wedeen VJ, Wald LL.
\newblock {Blipped-controlled aliasing in parallel imaging for simultaneous
  multislice echo planar imaging with reduced g-factor penalty}.
\newblock Magn Res Med. 2011;67(5):1210--1224.

\bibitem{dwi-esp}
Galinsky VL, Frank LR.
\newblock {{S}imultaneous multi-scale diffusion estimation and tractography
  guided by entropy spectrum pathways}.
\newblock IEEE Trans Med Imaging. 2015;34(5):1177--1193.

\bibitem{Glover:2000}
Glover GH, Li TQ, Ress D.
\newblock {Image-based method for retrospective correction of physiological
  motion effects in fMRI: RETROICOR}.
\newblock Magn Reson Med. 2000;44(1):162--167.

\bibitem{Chang:2009B}
Chang C, Glover GH.
\newblock {Effects of model-based physiological noise correction on default
  mode network anti-correlations and correlations}.
\newblock Neuroimage. 2009;47(4):1448--1459.

\bibitem{2016JPhA...49M5001F}
{Frank} LR, {Galinsky} VL.
\newblock {Detecting spatio-temporal modes in multivariate data by entropy
  field decomposition}.
\newblock Journal of Physics A Mathematical General. 2016;49:395001.
\newblock doi:{10.1088/1751-8113/49/39/395001}.

\bibitem{fmri-efd}
{Frank} LR, {Galinsky} VL.
\newblock {{D}ynamic {M}ultiscale {M}odes of {R}esting {S}tate {B}rain
  {A}ctivity {D}etected by {E}ntropy {F}ield {D}ecomposition}.
\newblock Neural Comput. 2016;28(9):1769--1811.

\end{thebibliography}

\end{document}